\documentclass{article}
\usepackage[preprint]{neurips_2026}
\setcitestyle{numbers,square,comma}
\usepackage[T1]{fontenc}
\usepackage{amsmath,amssymb}
\usepackage{graphicx}
\usepackage{booktabs,array,calc}
\usepackage{fvextra}
\usepackage{caption}
\usepackage{placeins}
\usepackage{microtype}
\usepackage[hidelinks]{hyperref}
\providecommand{\tightlist}{\setlength{\itemsep}{0pt}\setlength{\parskip}{0pt}}
\title{Fuse-All MoE Inference Kernel with Load Balancing on TPUs}
\author{Rupeng Liu \quad Bangsheng Tang \quad Zongwei Zhou \\
\textbf{Chengji Yao \quad Yejin Lee} \\
Meta SuperIntelligence Lab \\
{\small\texttt{rupengliu@meta.com, bangsheng@meta.com, zongwei@meta.com}} \\
{\small\texttt{chengjiyao@meta.com, yejinlee@meta.com}}}
\date{}
\begin{document}
\raggedbottom
\maketitle
\begin{abstract}
Mixture-of-Experts (MoE) architectures activate only a subset of experts for each token, allowing large language models (LLMs) to scale parameter capacity without a proportional increase in computation per token. However, intermediate data movement and uneven expert loads limit the efficiency of MoE inference on TPUs. To address these bottlenecks, we introduce Fuse-All MoE, a TPU kernel design that jointly optimizes expert computation, output communication, and load balancing. The kernel fuses token collection, expert projections, gated activation, and output writes, retaining expert computation intermediates on chip. Fixed token-slot destinations allow scheduling and expert placement to change while preserving routing semantics. In prefill-decode (PD) disaggregated serving, Fuse-All MoE balances EP16 prefill workloads through scheduling and expert redistribution, while symmetric EP8TP2 shares active expert computation between the two TensorCore devices within each chip during decode. Our experiments show that Fuse-All MoE reduces kernel runtime for a measured skewed EP16 prefill workload from 2.3\ensuremath{\times} to 1.09\ensuremath{\times} the same fixed reference time for a balanced workload. Across 32--512 decode tokens, symmetric EP8TP2 reduces kernel latency by up to 21.3\% relative to fused EP16. With accompanying serving stack optimizations, the integrated system improves serving throughput by up to 34.5\% over the Megablox EP16 baseline.
\end{abstract}

\section{Introduction}\label{introduction}

Mixture-of-Experts (MoE) models increase model capacity while activating only a subset of experts for each token {[}1-3{]}. Expert parallelism distributes these experts across devices, but also makes inference sensitive to where the router sends work. A device with many active experts can delay the layer even when other devices are lightly loaded. Efficient execution must therefore address computation, communication, and load imbalance together.

These costs change between prefill and decode. Large prefill batches offer substantial weight reuse, but skewed routing can concentrate computation on a few devices. Small decode batches provide less work over which to amortize weight reads and communication. In this regime, the number of active experts can matter more than the number of assigned tokens. Balancing token counts alone may leave the dominant cost unchanged.

We present Fuse-All MoE, a Pallas implementation \citep{ref05} that co-designs expert computation, output transport, and load management. Its central contribution is a fused compute-and-return pipeline whose token-slot output contract remains unchanged as tile order and expert execution placement change. Indexed input collection, both projections, gated activation, and direct writes execute in one expert launch, retaining intermediates and reusable weights in VMEM while overlapping computation with output communication. Figure 1 shows this boundary; routing and upstream input exchange remain separate.

The second part addresses uneven routing. Row bucketing avoids padded computation, hot-expert scheduling exposes weight-prefetch overlap, and destination interleaving reduces traffic bursts. Static placement, runtime chipmate redistribution, and preloaded hot-expert copies change where work executes. A symmetric TP2 decode variant shares each active expert across chipmates. The implemented cross-chip copy path targets decode; prefill integration is an extension.

\begin{figure}[htbp]
\centering
\includegraphics[width=.94\linewidth]{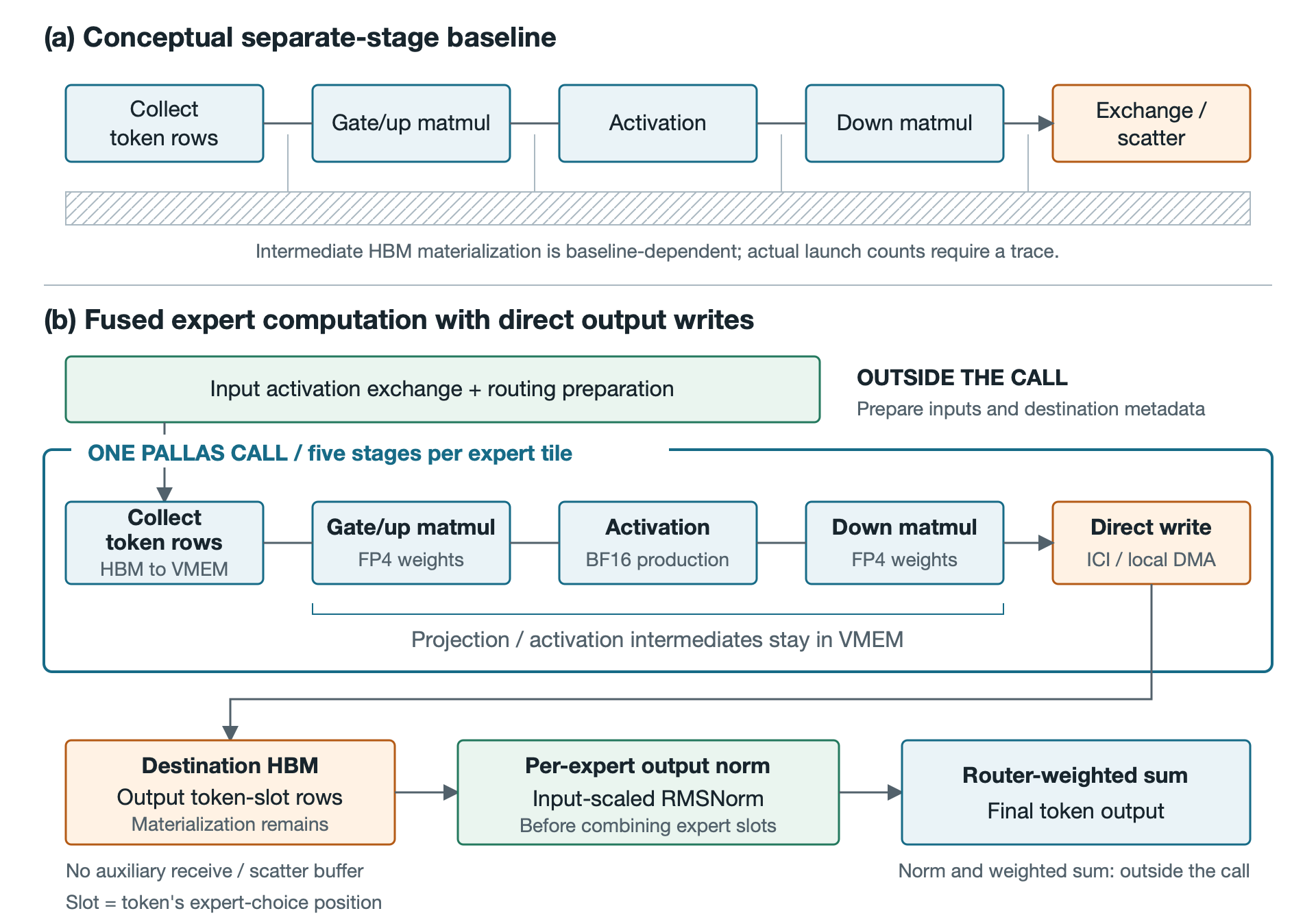}
\caption{ Fusion boundary. Gate/up and activated intermediates stay in VMEM; expert results go directly to final token-slot buffers. Routing, input exchange, and postprocessing remain outside the expert launch. Boxes show logical stages, not measured kernel counts.}
\label{fig:1}
\end{figure}

The paper develops two complementary contributions:

\begin{itemize}
\tightlist
\item
  \textbf{A fused EP16 compute-and-return pipeline.} We integrate token collection, expert computation, and final-destination output exchange under explicit buffer-lifetime and completion rules. This removes intermediate materialization and receive-side redistribution while preserving output ownership when work moves.
\item
  \textbf{Regime-specific balancing within that pipeline.} For PD serving, EP16 uses load-aware tile and traffic scheduling, placement, and runtime redistribution for prefill, while symmetric EP8TP2 shares active experts across chipmates with aligned tile computation for decode. Preloaded copies provide a complementary cross-chip placement mechanism. The designs preserve the router\textquotesingle s expert choices and a common token-slot output layout.
\end{itemize}

Section 3 develops the base EP16 kernel. Section 4 introduces imbalance-handling mechanisms and the decode variant. Sections 5 and 6 present evaluation and remaining limitations, distinguishing expert-kernel latency, full-layer cost, and serving throughput.

\section{Background and Problem Definition}\label{background-and-problem-definition}

\subsection{TPU Hardware and the Unit of Parallelism}\label{tpu-hardware-and-the-unit-of-parallelism}

A TPU7x (Ironwood) chip exposes two TensorCore devices to JAX, each with its own HBM address space. The chiplets communicate over a die-to-die link; physical chips use the inter-chip interconnect (ICI). A rank denotes one EP device: the base EP16 configuration spans sixteen devices on eight chips. A topology map identifies chipmates. Section 4.6 distinguishes this layout from EP8 x TP2.

Each TensorCore contains a matrix unit (MXU), a vector processing unit (VPU), and scalar control. TPU7x\textquotesingle s MXU uses a 256\ensuremath{\times}256 systolic array {[}4{]}. VMEM holds vector tiles and intermediates; SMEM holds scalar metadata. SparseCores have separate irregular-access hardware and memory. Below, K denotes the matmul reduction dimension and N its output width.

\begin{table}[!htbp]
\centering
\small
\renewcommand{\arraystretch}{1.12}
\caption{ TPU7x specifications relevant to this design {[}4, 9, 10{]}. Compute and HBM bandwidth are per physical chip, unless stated otherwise.}\label{tab:1}
\begin{tabular}{@{}
  >{\raggedright\arraybackslash}p{(\linewidth - 4\tabcolsep) * \real{0.2300}}
  >{\raggedright\arraybackslash}p{(\linewidth - 4\tabcolsep) * \real{0.2700}}
  >{\raggedright\arraybackslash}p{(\linewidth - 4\tabcolsep) * \real{0.5000}}@{}}
\toprule
\begin{minipage}[b]{\linewidth}\raggedright
Resource
\end{minipage} & \begin{minipage}[b]{\linewidth}\raggedright
Specification
\end{minipage} & \begin{minipage}[b]{\linewidth}\raggedright
Consequence
\end{minipage} \\
\midrule
TensorCore devices / SparseCores & 2 / 4 per chip & EP device count and physical chip count differ. \\
VMEM & 64 MiB per TensorCore & Weights, intermediates, and DMA staging compete for capacity. \\
HBM & 192 GiB capacity; approximately 7.38 TB/s & Use chip-level bytes and chip-level peak throughput consistently. \\
Dense compute & 2,307 TFLOPs/s BF16; 4,614 TFLOPs/s FP8 & A roofline must match the arithmetic mode actually used. \\
ICI & 1,200 GB/s bidirectional per chip & This is aggregate bandwidth, not a per-destination guarantee. \\
Intra-chip link & Documented as six times a 1-D ICI link & Chipmate weight transfers avoid inter-chip torus traffic. \\
\bottomrule
\end{tabular}
\end{table}

The peak compute-to-HBM ratios are approximately 313 FLOPs/byte for BF16 (2307e12 / 7.38e12) and 625 for FP8. These roofline ridge points bound the intensity needed to reach peak compute. Actual mixed FP8\ensuremath{\times}FP4 throughput also depends on generated instructions, MXU utilization, and achieved bandwidth.

\begin{figure}[htbp]
\centering
\includegraphics[width=\linewidth]{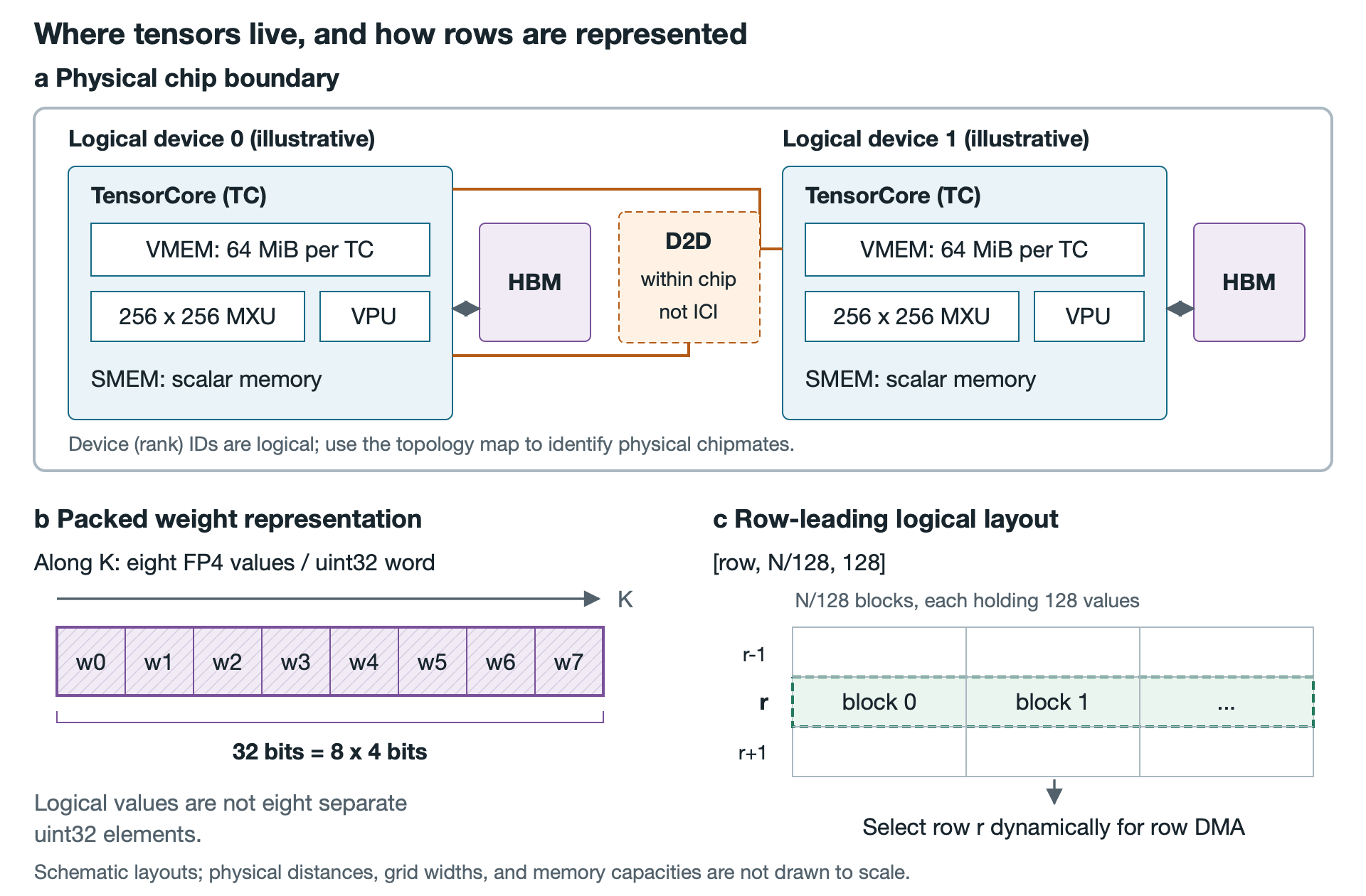}
\caption{ Hardware and representation. Chipmates have separate address spaces; scalar metadata and vector tiles use separate memories. FP4 packing reduces weight storage, and a leading row axis supports row-addressed DMA.}
\label{fig:2}
\end{figure}

\subsection{Pallas, Layout, and Software Pipelines}\label{pallas-layout-and-software-pipelines}

Pallas expresses tiled kernels that Mosaic compiles into statically scheduled TPU programs. Fuse-All MoE explicitly issues local and remote DMAs ahead of use and waits on completion semaphores. JAX Megablox provides a public Pallas grouped-matmul implementation {[}15{]}.

HBM fetches, MXU/VPU computation, and output writes can overlap subject to buffer dependencies. Shared bandwidth, scalar descriptor issue, and interconnect contention can still expose stalls.

Activations and outputs have a leading token-row axis and vector-sized trailing dimensions, such as {[}rows, N/128, 128{]}. Weights pack eight FP4 values per 32-bit word along K. Logical matrices, packed storage, and MXU tiles are distinct representations; packing does not determine arithmetic precision.

\subsection{MoE Semantics and Notation}\label{moe-semantics-and-notation}

A routed assignment is a (token, selected-expert) pair; its slot is the position among the token\textquotesingle s \(k\) choices, not a global expert ID. W1 combines gate/up weights; W2 holds down-projection weights. Token collection denotes indexed reads from the input table into VMEM, not inter-device input exchange.

\begin{table}[!htbp]
\centering
\small
\renewcommand{\arraystretch}{1.12}
\caption{ Notation. Unless explicitly stated otherwise, token and routed-row counts are global across the EP mesh.}\label{tab:2}
\begin{tabular}{@{}
  >{\raggedright\arraybackslash}p{(\linewidth - 2\tabcolsep) * \real{0.2300}}
  >{\raggedright\arraybackslash}p{(\linewidth - 2\tabcolsep) * \real{0.7700}}@{}}
\toprule
\begin{minipage}[b]{\linewidth}\raggedright
Symbol
\end{minipage} & \begin{minipage}[b]{\linewidth}\raggedright
Meaning
\end{minipage} \\
\midrule
\(T\), \(T_{\mathrm{valid}}\) & Token count of the compiled tensor shape, including padding, and the valid-token count used to plan work. \\
\(k\), \(M\) & Experts selected per token and total (token, expert-choice) assignments, M = Tk for complete routing. \\
\(E\), \(P\) & Total experts and EP devices/ranks. Initial expert assignment has \(E/P\) experts per rank. \\
\(d\), \(f\) & Hidden width and expert intermediate width. \\
\(c\), \(s\) & Owned tokens per rank, \(c\) = \(T/P\), and routing slot \(s\) in {[}0, \(k\)). \\
\(r_{e}\), \(t_{m}\) & Rows assigned to expert e and the maximum rows in one compute tile for that expert. \\
\(q\), \(g_{e}\) & Quantization-block width along K, and an expert\textquotesingle s per-projection global scale. \\
\bottomrule
\end{tabular}
\end{table}

For token \(t\), let \(e(t,s)\) denote the expert in slot \(s\) and \(a_{t,s}\) its router weight. A standard SiLU- or GELU-gated expert has \(d\times f\) gate/up matrices and an \(f\times d\) down matrix:

\[
\begin{aligned}
h_e(x) &= \operatorname{act}(xW_{\mathrm{gate},e}) \odot (xW_{\mathrm{up},e}),\\
z_e(x) &= h_e(x)W_{\mathrm{down},e},\\
y_t &= \sum_s a_{t,s}\,z_{e(t,s)}(x_t).
\end{aligned}
\]

This example activates the gate before multiplication by the up vector; variants that also transform the up branch require a matching fused activation. Routing remains outside the kernel, and per-expert normalization precedes the weighted sum (\S{}3.1).

Expert placement determines the producer; token ownership determines the destination. For contiguous token partitions:

\begin{Verbatim}[fontsize=\footnotesize,breaklines=true,breakanywhere=true]
destination(t) = floor(t / c)
local_token(t) = t mod c
output_row(t,s) = local_token(t) * k + s
\end{Verbatim}

Rebalancing changes the producer without changing these destinations, allowing direct writes to compose with load balancing.

\begin{figure}[htbp]
\centering
\includegraphics[width=.94\linewidth]{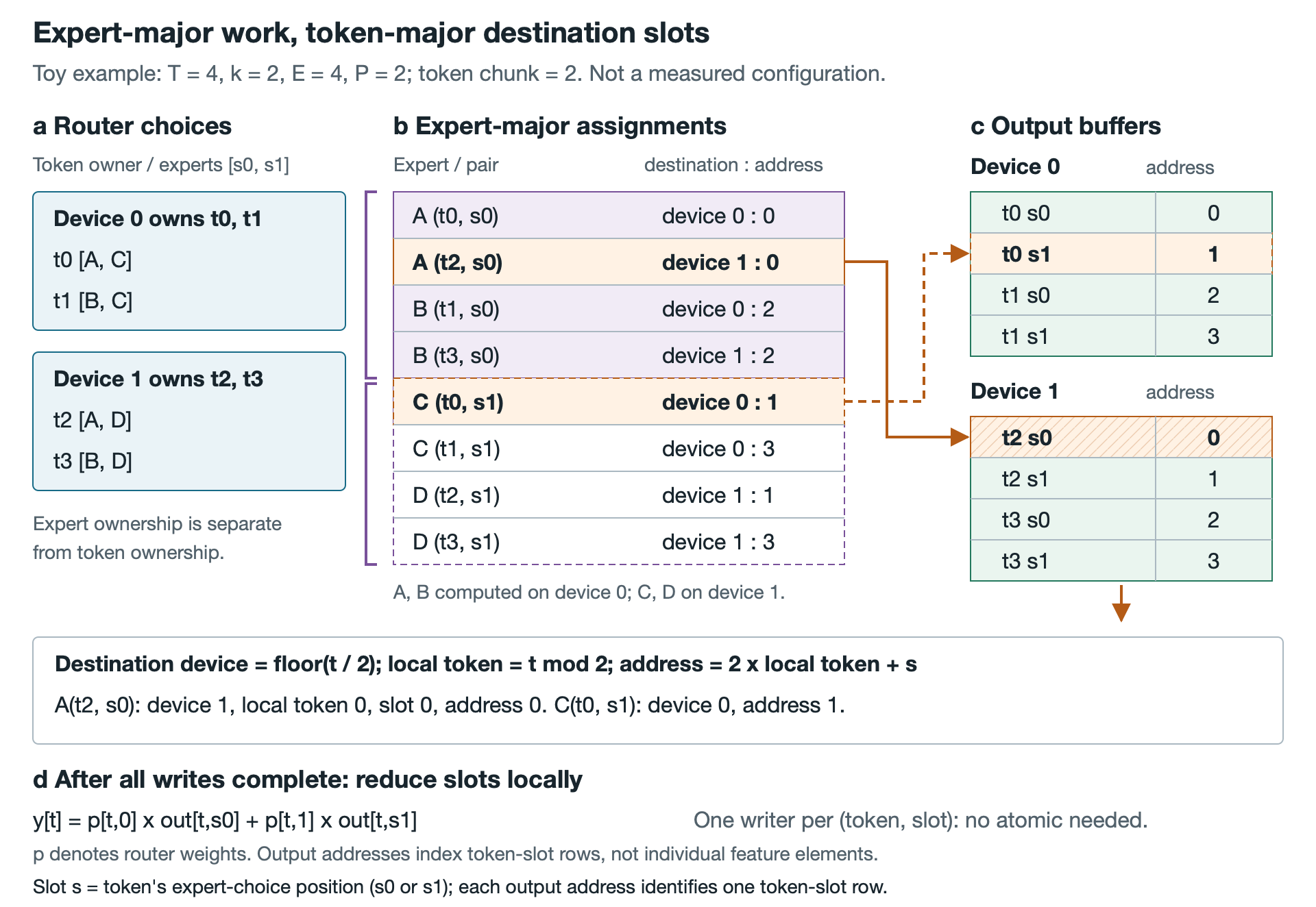}
\caption{ Routing example: \(T=4\), \(E=4\), \(P=2\), \(k=2\). Rank 0 owns A/B and rank 1 owns C/D; tokens 0--3 choose {[}A,C{]}, {[}B,C{]}, {[}A,D{]}, and {[}B,D{]}. A\textquotesingle s result for token 2, slot 0 returns to rank 1, row 0; C\textquotesingle s result for token 0, slot 1 returns to rank 0, row 1. Expert sorting changes compute order, not destinations. This is not a benchmark workload.}
\label{fig:3}
\end{figure}

\subsection{Baseline and Serving Boundary}\label{baseline-and-serving-boundary}

The baseline uses the Megablox grouped-matrix-multiplication kernels for expert computation under EP16 on a 2x2x2 TPUv7 slice, comprising eight physical chips and sixteen TensorCore devices. It does not use our fused compute-and-return kernel. Full-layer comparisons include routing, activation, exchange, and combination. Weight format, activation quantization, expert normalization, routes, and output semantics must be matched; exact baseline launch and HBM-transfer counts require compiled-trace confirmation.

The model runner sorts assignments by expert and replicates the full input-row table in every device\textquotesingle s HBM before the expert launch. Sequence-parallel inputs are all-gathered first; prequantized FP8 inputs include scales. Token collection then reads selected rows locally, including for borrowed experts without additional token exchange. The launch returns one row per token-slot; postprocessing forms the weighted sum.

\subsection{Related Work}\label{related-work}

\textbf{Models and routing.} GShard \citep{ref02} demonstrates large-scale TPU MoE training; Switch Transformers \citep{ref03} studies simplified routing and stable low-precision training. Expert Choice \citep{ref13} balances load through fixed per-expert token buckets, allowing variable expert counts per token. Fuse-All MoE instead preserves token-choice top-\(k\) assignments.

\textbf{Expert computation.} MegaBlocks \citep{ref11} uses block-sparse GPU training to handle unequal expert loads without token dropping, while retaining block-granularity padding. Tutel \citep{ref12} optimizes GPU dispatch/combine and adaptive MoE execution. On TPUs, JAX Megablox \citep{ref15} provides Pallas grouped matmuls, integrated with separate routing and gating in MaxText {[}16{]}. MegaBlocks and Megablox are distinct references.

\textbf{Communication and serving.} DeepEP \citep{ref14} specializes GPU expert-parallel dispatch and combine. SGLang \citep{ref06} and vLLM \citep{ref07} address runtime execution and attention-cache management. Fuse-All MoE targets the TPU expert launch: token collection, both projections, activation, and addressed output writes. Routing, upstream exchange, and the router-weighted sum remain separate.

Relative to the standalone TPU grouped-matmul building blocks in Megablox and MaxText {[}15, 16{]}, our contribution couples the full expert-compute-and-return pipeline with load-aware scheduling and execution placement. The stable output contract connects these mechanisms: expert producers may change while final token-slot destinations remain fixed. We preserve token-choice routing rather than introducing a new routing algorithm.

\FloatBarrier
\section{Base EP16 Fused Kernel Design}\label{base-ep16-fused-kernel-design}

We first describe the EP16 kernel: experts are partitioned across sixteen TensorCore devices on eight chips, and each expert is computed entirely by its owning device. Each device runs a fused tile pipeline that writes results directly to token-slot destinations. This section establishes the dataflow, precision, memory reuse, and communication contract for fixed expert ownership. Section 4 adds load-aware scheduling and placement, including the EP8 x TP2 decode variant.

\subsection{Fusion and the Output Contract}\label{fusion-and-the-output-contract}

Each device consumes activations, expert weights, row counts, token indices, and output destinations to produce one row per routed assignment:

\begin{enumerate}
\tightlist
\item
  \textbf{Collect token rows.} Indexed row DMAs gather inputs directly from HBM into VMEM, avoiding an expert-sorted activation tensor in HBM.
\item
  \textbf{Gate/up matmul.} The kernel unpacks W1, applies scales, and computes gate and up vectors together.
\item
  \textbf{Gated activation.} The standard \texttt{act(gate)\ *\ up} operation writes directly to the down projection\textquotesingle s VMEM input, removing a separate activation kernel and intermediate copy.
\item
  \textbf{Down-projection matmul.} Multiplication by W2 produces a hidden-width result, staged for output DMA while later tiles compute.
\item
  \textbf{Direct write.} Local or remote DMA writes each result to its final token-slot row, removing auxiliary receive storage and receive-side scatter-add redistribution.
\end{enumerate}

Each device\textquotesingle s output buffer holds \(ck\) rows, one per token-slot. Transport writes but does not sum these rows, unlike conventional reduce-scatter. Separate postprocessing converts to FP32, combines router-weighted results, and casts to the requested dtype. For RMS-normalized expert blocks, scaling and normalization precede the sum:

\begin{Verbatim}[fontsize=\footnotesize,breaklines=true,breakanywhere=true]
u[t,s] = (1 + norm_weight) * z[t,s]             # scale expert result
v[t,s] = u[t,s] / sqrt(mean(u[t,s]^2) + 1e-8)  # normalize each result
y[t]   = sum_s a[t,s] * v[t,s]                 # combine expert choices
\end{Verbatim}

RMSNorm of the weighted sum is not equivalent to this per-expert normalization. Postprocessing consumes the tiled slots directly; its logical reshape need not materialize an HBM transpose.

\begin{table}[!htbp]
\centering
\small
\renewcommand{\arraystretch}{1.12}
\caption{ What fusion removes and what it retains.}\label{tab:3}
\begin{tabular}{@{}
  >{\raggedright\arraybackslash}p{(\linewidth - 2\tabcolsep) * \real{0.4000}}
  >{\raggedright\arraybackslash}p{(\linewidth - 2\tabcolsep) * \real{0.6000}}@{}}
\toprule
\begin{minipage}[b]{\linewidth}\raggedright
Boundary or storage
\end{minipage} & \begin{minipage}[b]{\linewidth}\raggedright
Fused path
\end{minipage} \\
\midrule
Grouped activation materialization & Removed: token collection goes directly into VMEM. \\
gate/up matmul / activation / down-projection matmul intermediates in HBM & Removed within the fused expert-compute path. \\
Auxiliary receive-and-scatter buffer & Removed: writes target final token-slot addresses. \\
Final expert-slot output & Retained; the main HBM-output path materializes it before postprocessing. \\
Routing preparation and upstream activation exchange & Outside this fused launch; include their cost in full-layer measurements. \\
Weighted sum and per-expert output normalization & Outside this launch, implemented by a Pallas postprocessing kernel or JAX path. \\
\bottomrule
\end{tabular}
\end{table}

\subsection{Low-Precision Representation and Arithmetic}\label{low-precision-representation-and-arithmetic}

\subsubsection{Packed weights and scales}\label{packed-weights-and-scales}

Weights use FP4 e2m1, packing eight values per uint32 word along K and reinterpreting them without a weight copy. FP32 block scales have shape {[}experts, K\_blocks, 1, N{]}; each scales a group of K values. A separate global scale applies to each expert\textquotesingle s gate, up, or down projection.

Block width is recovered from the scale shape under a power-of-two policy, with trailing blocks padded alongside weights. It affects scaling and tile selection and must be reported with the weight format.

\subsubsection{Where scaling occurs}\label{where-scaling-occurs}

Weight scales follow each block matmul when their width is an integer multiple of the inner K step; otherwise they are folded into BF16 weight slices before multiplication. This is independent of activation precision: W4A8 uses eight-bit activations, W4A16 BF16. At K=4096, compatible 256/512-wide blocks postscale in either mode; 128-wide blocks prescale.

W4A8 quantizes each activation row by K-block absmax, handling zero blocks without division by zero. Prequantized FP8 inputs carry scales and bypass input quantization; the newly computed down-projection input still requires precision conversion.

\begin{figure}[htbp]
\centering
\includegraphics[width=\linewidth]{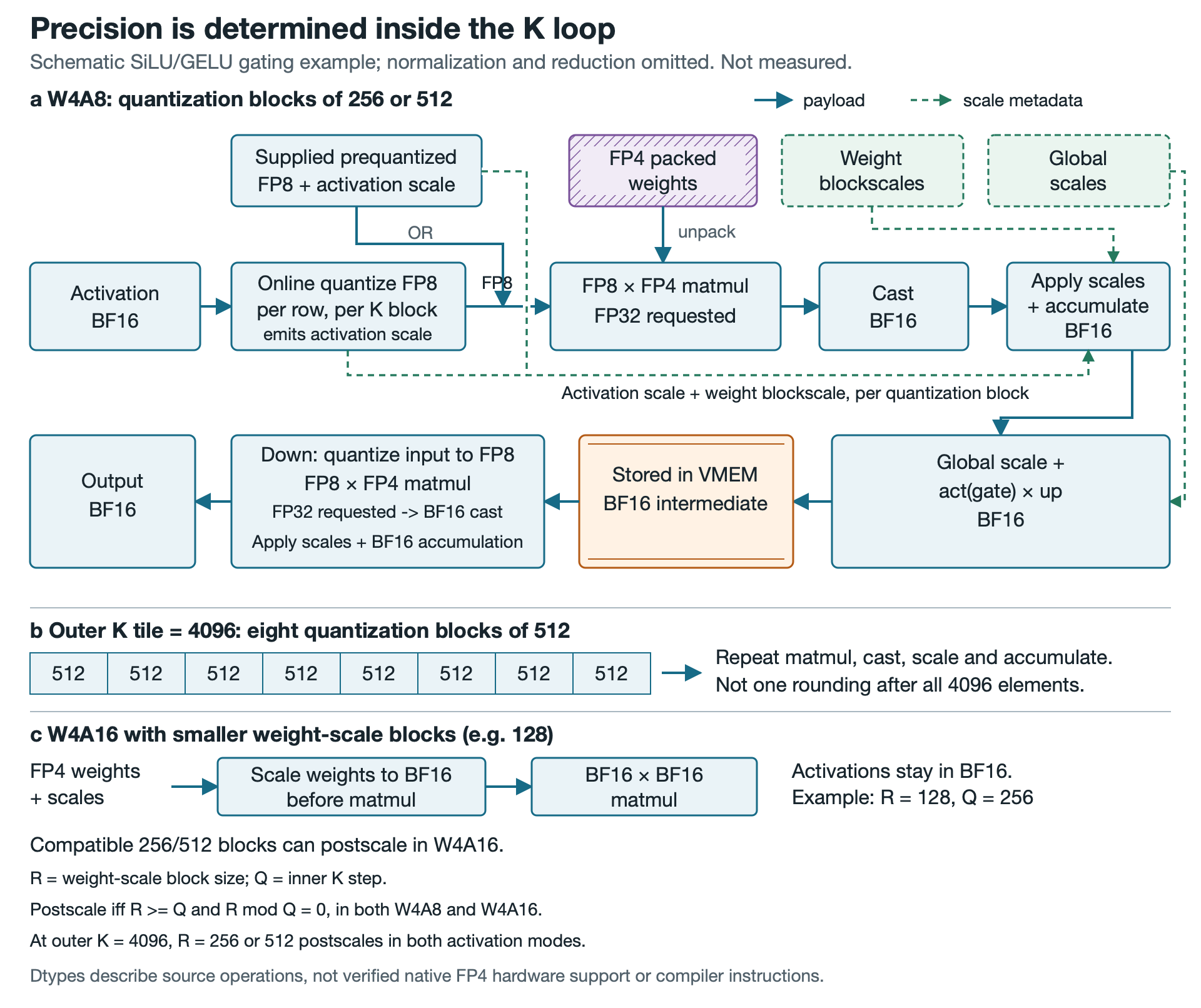}
\caption{ Precision boundaries for standard gated activation. FP4 block matmuls request FP32 results, cast to BF16 before scaling and summation. The K=4096 outer tile contains eight 512-wide quantization blocks; the down input is converted to its matmul precision. This shows source-level dataflow, not native instructions or measured error.}
\label{fig:4}
\end{figure}

\subsubsection{Accumulation Precision}\label{accumulation-precision}

Quantization-block matmuls request FP32 results but cast them into a BF16 shared accumulator in the BF16-output FP4 path. Block scaling, summation, global scaling, activation, and stored intermediates therefore do not remain FP32 throughout.

BF16 accumulation saves 16 KiB per row for an 8192-column accumulator relative to FP32, less for narrower N tiles. An outer K tile may contain multiple conversion-and-accumulation steps. Layer- and model-level accuracy validation is therefore essential, especially for long K or multiple outer K tiles.

Global scales apply after each projection\textquotesingle s final K tile, separately for gate and up columns. Completing all K reductions for one output-column chunk before the next exposes scaling and stores for overlap with later matmuls.

\subsection{Software Pipeline and Weight Residency}\label{software-pipeline-and-weight-residency}

\subsubsection{Buffers and dependencies}\label{buffers-and-dependencies}

The outer loop processes expert tiles of at most \(t_{m}\) rows, including alignment padding. Each iteration pipelines next-input collection and weight prefetch with current computation and output writes. Expert identity and weight-buffer capacity determine reuse.

\begin{table}[!htbp]
\centering
\small
\renewcommand{\arraystretch}{1.12}
\caption{ VMEM data buffers and SMEM destination metadata. Ring depths and tile shapes share a VMEM budget.}\label{tab:4}
\begin{tabular}{@{}
  >{\raggedright\arraybackslash}p{(\linewidth - 4\tabcolsep) * \real{0.3100}}
  >{\raggedright\arraybackslash}p{(\linewidth - 4\tabcolsep) * \real{0.2000}}
  >{\raggedright\arraybackslash}p{(\linewidth - 4\tabcolsep) * \real{0.4900}}@{}}
\toprule
\begin{minipage}[b]{\linewidth}\raggedright
Buffer
\end{minipage} & \begin{minipage}[b]{\linewidth}\raggedright
Multiplicity
\end{minipage} & \begin{minipage}[b]{\linewidth}\raggedright
Lifetime or purpose
\end{minipage} \\
\midrule
Collected input rows & 2 tiles & Current compute and next-tile token collection use different slots. \\
W1 and its block scales & Circular weight buffers & Current-expert consumption, cross-tile prefetch, and conditional residency. \\
W2 and its block scales & Tuned ring & Issued early enough to overlap token collection/gate/up matmul; repeated-expert hits skip loads. \\
Activated intermediate & 1 tile & Written by gate/up matmul\textquotesingle s activation and consumed by down-projection matmul. \\
Shared accumulator / down-projection matmul tile & Shared scratch / 1 output tile & Computation and conversion before temporary send buffers. \\
Temporary send buffers & 3 tiles & Rows remain stable until local/remote source DMA completion. \\
Output destination metadata & 2 bounded SMEM arrays & Destination rank and output-row index for the active window. \\
\bottomrule
\end{tabular}
\end{table}

\begin{figure}[htbp]
\centering
\includegraphics[width=\linewidth]{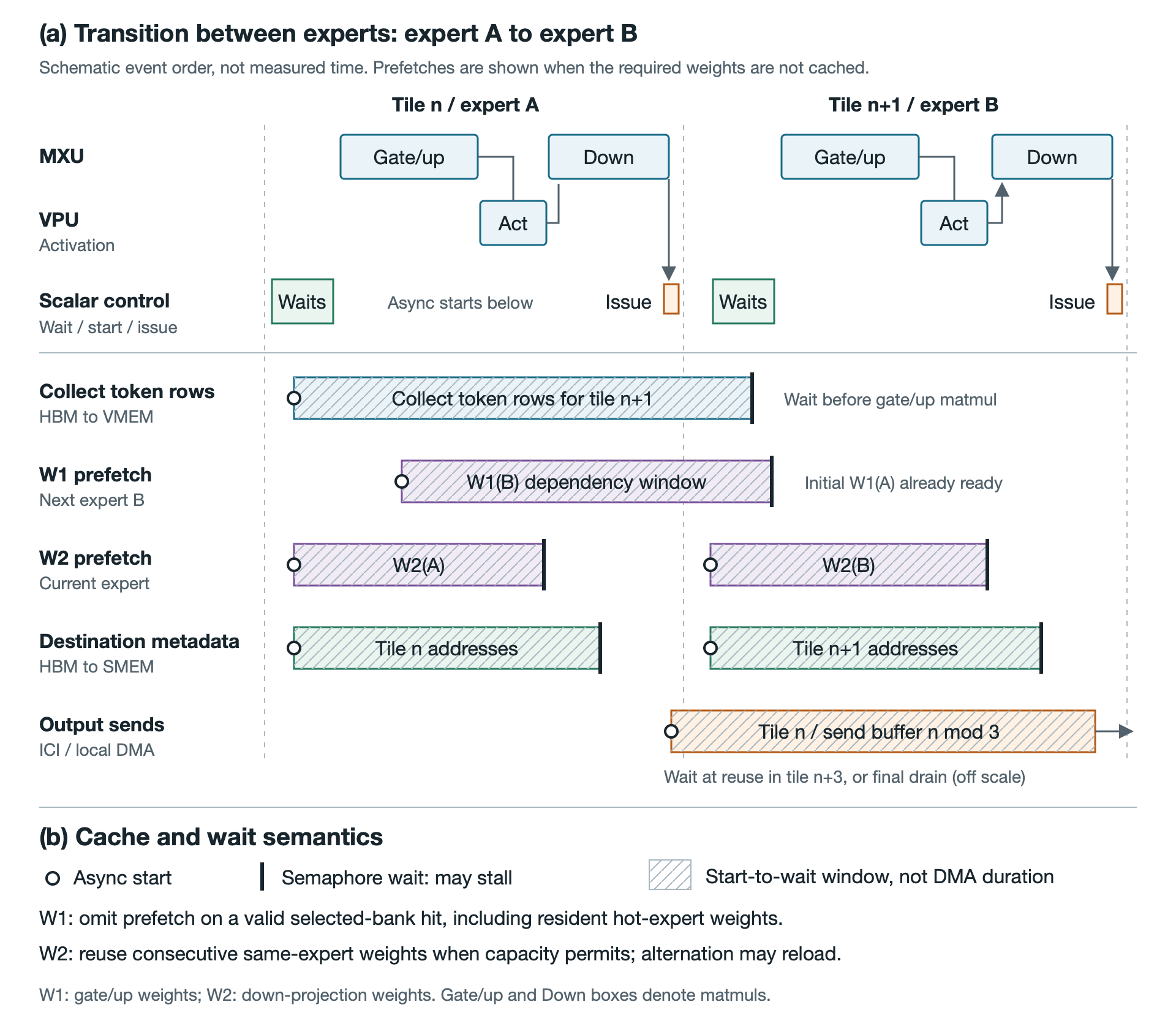}
\caption{ Pipeline dependencies. Start-to-wait intervals show available overlap, not measured DMA durations {[}4{]}. Same-expert hits omit weight transfers and waits; scalar issue and late transfers may remain exposed.}
\label{fig:5}
\end{figure}

Before reusing a send slot, the kernel drains its outstanding DMAs. It then starts destination-metadata and next-input fetches. Down weights are requested before the current-input wait; gate/up weights are usually prefetched by the preceding tile.

Prefetch may overwrite a weight buffer only after its last consumer. A separate bank or sufficient remaining capacity permits early next-expert prefetch. Down weights follow the same wait-and-refill discipline. Completed rows enter a send buffer and leave once destination metadata is ready. Full-matrix residency requires a ring large enough to retain every weight tile.

\subsubsection{Same-expert caching}\label{same-expert-caching}

Let \(S_{1}\) and \(S_{2}\) count the gate/up and down weight tiles. A ring holding an entire matrix lets consecutive tiles of that expert skip loads and waits; a smaller ring overwrites earlier tiles. For \(L\) consecutive input tiles and packed expert size \(B_{\mathrm{expert}}\), excluding scales:

\begin{Verbatim}[fontsize=\footnotesize,breaklines=true,breakanywhere=true]
cacheable_i = (weight_ring_depth_i >= S_i)
weight bytes for L consecutive tiles:
    without reuse: L * B_expert
    full residency: B_expert
\end{Verbatim}

With both projections resident, \(L=16\) cuts this run\textquotesingle s weight reads 16-fold, not total layer traffic. Activations, outputs, scales, and communication remain. The example uses 16 MiB of packed W1, 8 MiB of W2, and 0.375 MiB of FP32 scales at block width 512.

\begin{figure}[htbp]
\centering
\includegraphics[width=.97\linewidth]{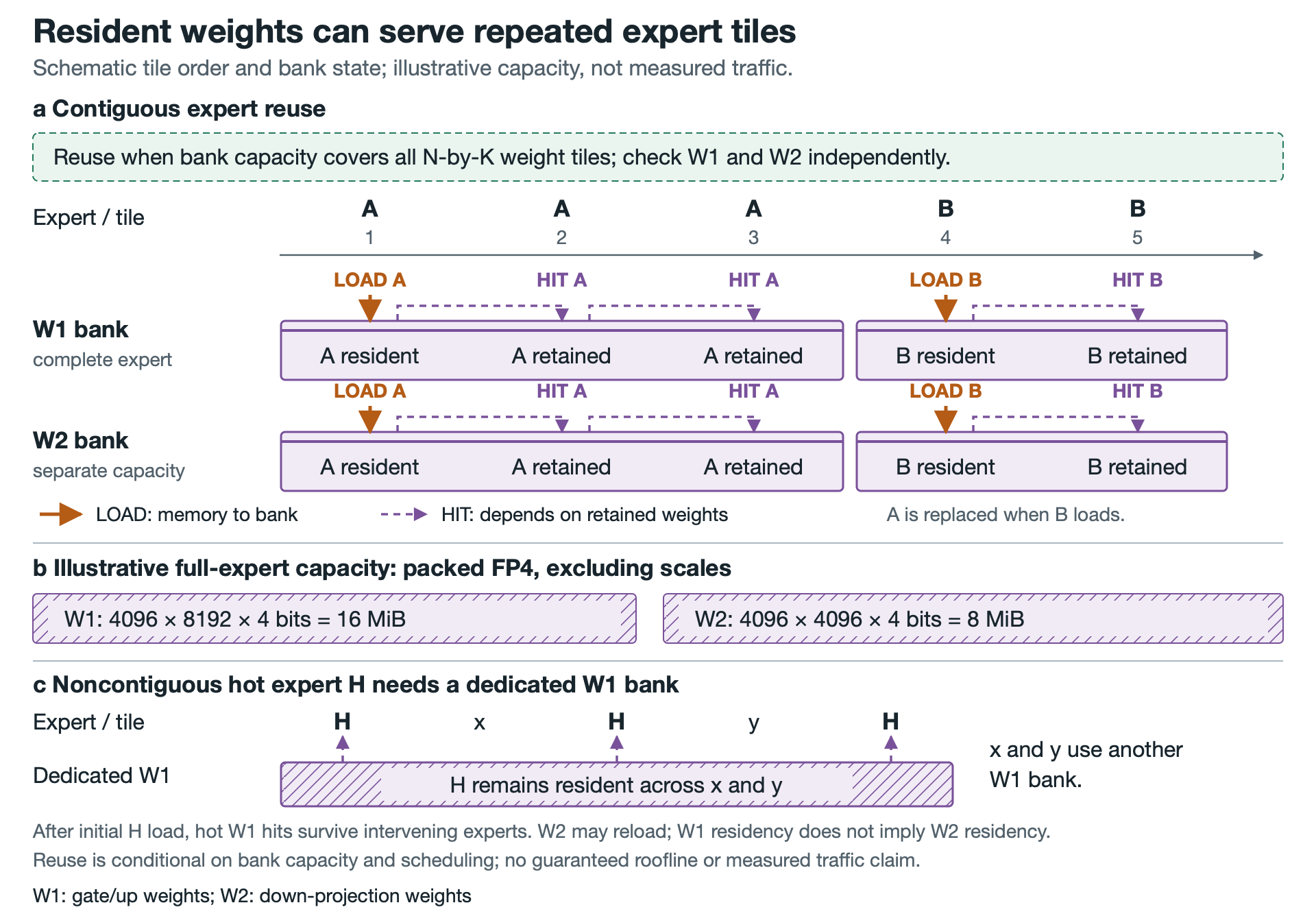}
\caption{ Consecutive tiles retain W1/W2 when both rings fit. The hot-expert extension in Section 4.2 reserves a separate W1 bank. A hit removes that projection\textquotesingle s reload, not all tile traffic.}
\label{fig:6}
\end{figure}

\subsubsection{Arithmetic intensity and its limits}\label{arithmetic-intensity-and-its-limits}

A gated expert performs approximately 6 * r\_e * d * f matmul FLOPs and stores 3 * d * f / 2 packed weight bytes, excluding scales. One load gives weight-only intensity near 4 * r\_e FLOPs/byte; per-tile reloads amortize only over tile rows.

At \(T=8192\), \(k=8\), \(E=256\), there are 65,536 global assignments: 256 rows per expert and 4096 per EP16 rank under balanced routing. Skew and alignment determine actual group lengths and tile counts.

Compute-bound claims must account for actual HBM traffic, achieved mixed-precision throughput, padding, scalar work, and communication. Resource work divided by achievable rate gives lower bounds; approaching their maximum requires sufficient buffering and overlap.

\subsection{Direct-Write Exchange and Synchronization}\label{direct-write-exchange-and-synchronization}

\subsubsection{Unique Output Ownership}\label{unique-output-ownership}

If each (global token, slot) appears once, exactly one producer writes each output row, even when two slots choose the same expert. The later router-weighted sum combines expert results {[}1, 11{]}. Direct writes eliminate auxiliary receive-and-scatter storage, not the final landing buffer.

The wrapper skips output zero-fill only for complete routing with even token sharding. Invalid or padded routes map to a real expert with zero router weight, preserving coverage. Callers that drop assignments must initialize unwritten rows; buffer shape does not establish that a custom route list is a permutation.

\subsubsection{Destination-Aware DMA Issue}\label{destination-aware-dma-issue}

The generic path issues complementary local/remote descriptors per valid row, with sizes (1,0) or (0,1), avoiding a destination-dependent branch. Zero-size descriptors carry no payload. Per-buffer counts track local and remote completions.

Large destination-sorted tiles instead use contiguous destination ranges and compile-time device addresses, issuing one applicable descriptor per row. This reduces address-encoding work but does not apply to interleaved destinations.

\subsubsection{Output Staging}\label{output-staging}

Three rotating send buffers leave two intervening tiles before reuse, versus one with two buffers. They are transport storage, distinct from the permanent \(k\) expert-choice slots. Greater depth buys overlap slack at a VMEM cost.

\begin{figure}[htbp]
\centering
\includegraphics[width=.97\linewidth]{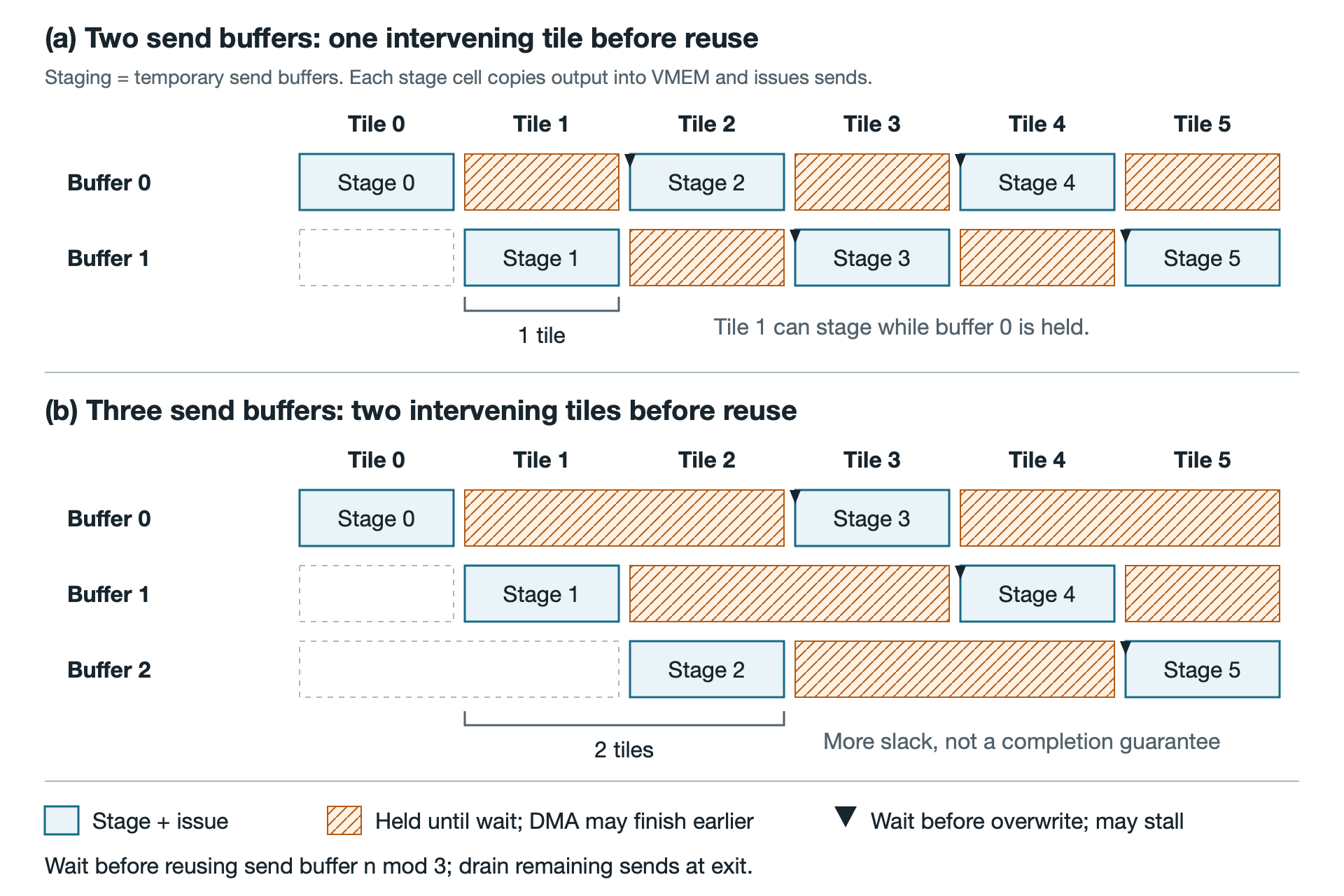}
\caption{ Send-buffer lifetime. Each buffer is reused every third iteration only after its outgoing DMAs complete; late transfers safely delay reuse.}
\label{fig:7}
\end{figure}

\subsubsection{Synchronization}\label{synchronization}

\begin{enumerate}
\tightlist
\item
  \textbf{Entry readiness.} Each rank completes any zero-fill and signals entry. The wait occurs before tile 0\textquotesingle s first remote activity, overlapping initial compute. Even without zero-fill, the peer must enter before its output or receive semaphore is addressed.
\item
  \textbf{Source lifetime.} Send-buffer reuse waits for recorded local-copy and remote-send completions, then clears the counters. Source completion does not establish receiver-wide completion.
\item
  \textbf{Tail completion.} Exit drains output staging. Ranks without compute still perform the entry wait.
\item
  \textbf{Receive completion and exit.} An EP barrier follows local drains; each rank then drains expected receive completions before exposing outputs.
\end{enumerate}

\begin{figure}[htbp]
\centering
\includegraphics[width=\linewidth]{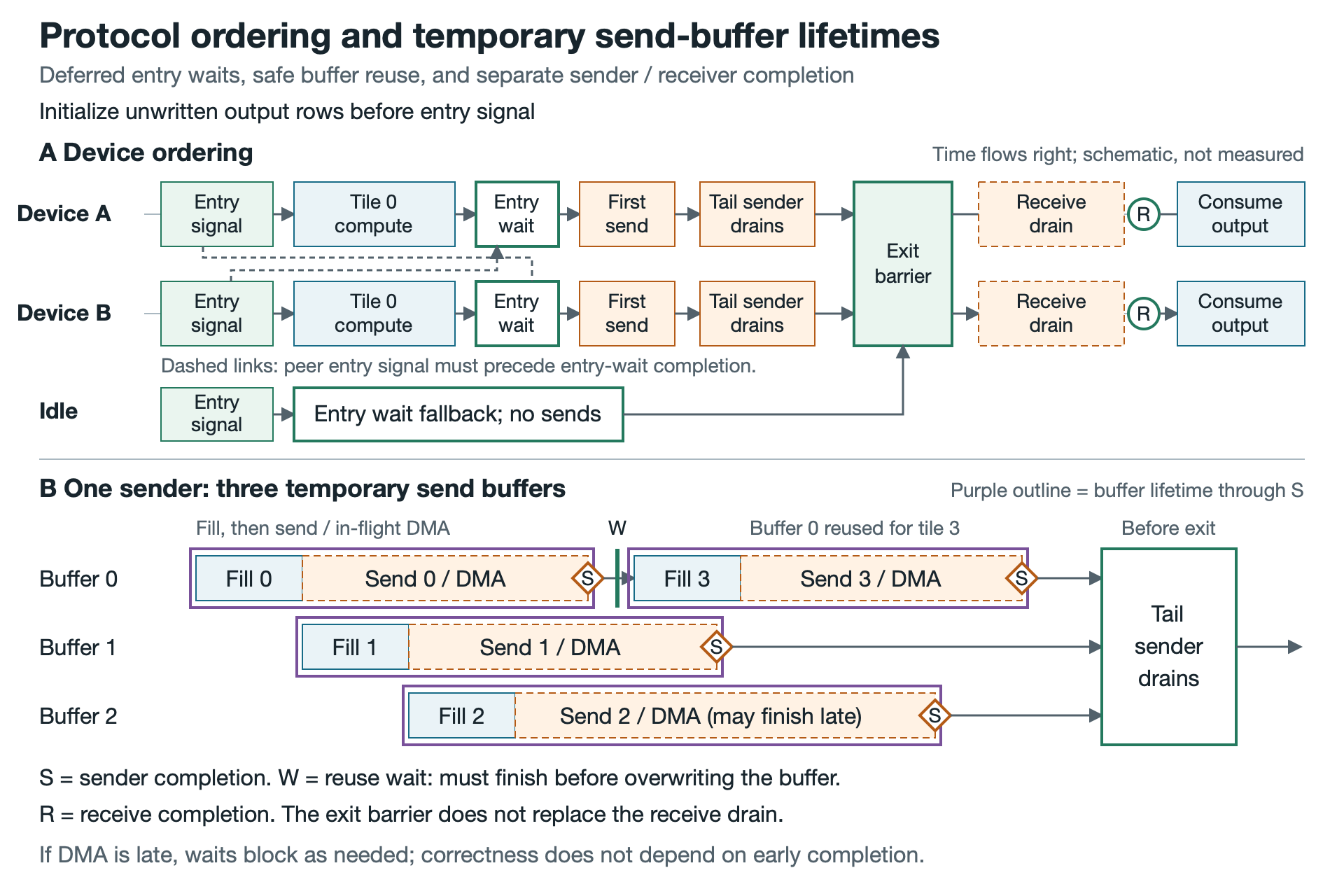}
\caption{ Entry readiness protects destinations, send drains protect sources, and receive completion protects consumers. Section 4.3 adds rebalancing-specific weight-transfer obligations. Safety does not depend on transfers fitting the overlap window.}
\label{fig:8}
\end{figure}

The loop has no periodic cross-rank barriers, but retains input, weight, metadata, and buffer-reuse waits. Early issue reduces exposed waits without weakening dependencies.

\subsubsection{Algorithm 1: base EP16 tile pipeline}\label{algorithm-1-base-ep16-tile-pipeline}

Dependency-level outline for fixed expert ownership. Asynchronous stages overlap subject to buffer lifetimes. Section 4 changes schedules and producers while preserving this output and completion contract.

\begin{Verbatim}[fontsize=\footnotesize,breaklines=true,breakanywhere=true]
prepare routes, destinations, counts, and the fixed-ownership tile schedule
derive expected remote rows, excluding local copies
initialize unwritten outputs when required; signal entry readiness
 
for each scheduled expert tile:
    drain prior sends before reusing output staging
    fetch destinations and gather current/upcoming inputs
    prefetch weights without overwriting live buffers; reuse resident weights
    await input and W1; compute gate/up and activation in VMEM
    await W2; compute the down projection into output staging
    await destinations and entry readiness before first remote activity
    direct-write valid rows to unique token-slot destinations
    record source-completion counts
 
complete entry readiness even with no local tiles
drain all final-write sources
rendezvous across EP ranks; drain expected incoming output rows
outside the launch: normalize expert results, then apply the router-weighted sum
\end{Verbatim}

\subsection{Routing Metadata and Bounded SMEM Windows}\label{routing-metadata-and-bounded-smem-windows}

\subsubsection{Routing Preparation}\label{routing-preparation}

Device-side JAX operations prepare expert-sorted assignments, token indices, per-expert counts, and output destinations. Full-layer timing includes this preparation.

Kernel setup uses replicated expert counts to build each device\textquotesingle s tile schedule without another collective. In the base EP16 path, the schedule follows fixed expert ownership; the planners in Section 4 modify that schedule when work moves.

Replicated counts determine the runtime loop bound, including alignment padding. Compilation fixes scratch capacities and tile shapes. A shared tile selector and caller/kernel check enforce consistency. Section 4.3 describes the bound adjustment required when a device receives redistributed work.

\subsubsection{Bounded Metadata Windows}\label{bounded-metadata-windows}

Full int32 destination and output-row arrays would occupy 8*M SMEM bytes. Instead, HBM stores them in 1024-entry slabs, and each tile loads a slab-aligned SMEM window addressed relative to its base.

\begin{Verbatim}[fontsize=\footnotesize,breaklines=true,breakanywhere=true]
slab_base = floor(tile_row_start / 1024) * 1024
window_slabs = min(available_slabs, floor((2046 + t_m) / 1024))
\end{Verbatim}

For 64--512-row tiles, sufficiently long inputs use two slabs per array: 2048 entries or 8 KiB, totaling 16 KiB of SMEM. Short inputs cap the window; HBM padding makes tail DMAs safe. Metadata fetch begins early in the tile and completes before send-address lookup.

\begin{figure}[htbp]
\centering
\includegraphics[width=.97\linewidth]{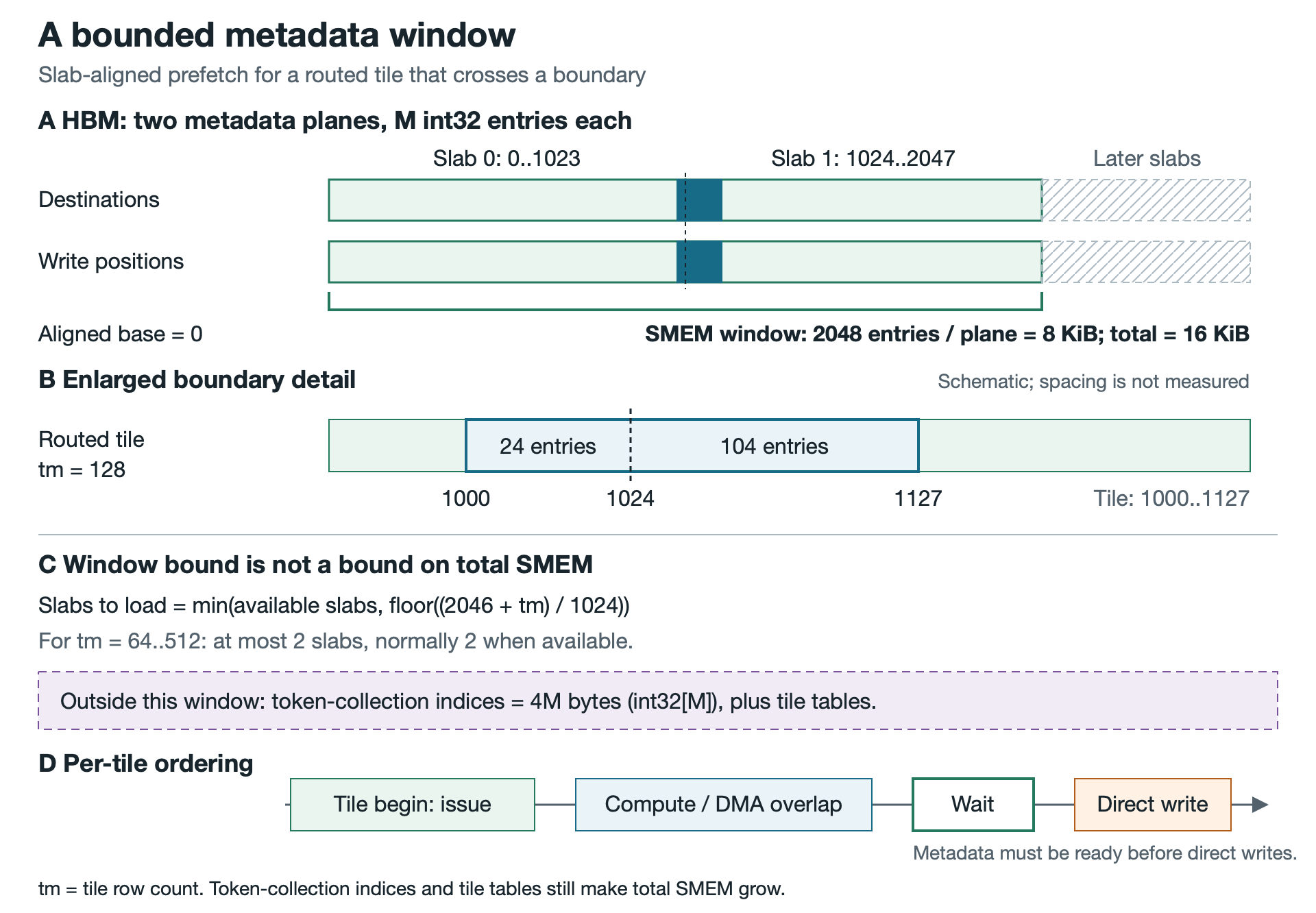}
\caption{ A tile starting at routed row 1000 crosses a slab boundary. The bounded two-array window supplies its send addresses; token indices and tile schedules remain workload-dependent.}
\label{fig:9}
\end{figure}

Token indices still cost about 4*M bytes, and tile schedules grow with workload size. At top-8, 32,768 global tokens require 262,144 indices, or 1 MiB before other metadata. Only the destination window is bounded for a fixed tile configuration.

\subsubsection{Receive accounting}\label{receive-accounting}

Expected remote-row counts exclude local copies. Under-counting leaves completion state outstanding; over-counting can wait for nonexistent rows. Section 4.3 extends this accounting when the producer changes.

Without a precomputed count, the kernel counts receives internally for \(M\leq 2048\) (\(T\leq 256\) at top-8); larger shapes use outer JAX preparation. This changes accounting location, not necessarily routing-operand VMEM residency. Its benefit must include preparation cost.

\subsection{Tile Selection and Static Shapes}\label{tile-selection-and-static-shapes}

Tile selection jointly chooses row height \(t_{m}\), K/N sizes for both matmuls, and both weight-ring depths under VMEM, alignment, spill-space, model-shape, and workload constraints. The caller uses the same row height to count tiles, checked at compilation.

Balanced and skewed routes can favor different tile shapes. Reproducible comparisons should report the selected shapes, precision mode, routing distribution, and source revision.

Static buckets avoid unexpected recompilation only when serving precompiles supported shape, dtype, and configuration combinations. Routing varies at runtime. Padding must preserve output coverage but be excluded from valid-load planning, so padded shape does not select an inappropriate policy.

\FloatBarrier
\section{MoE Imbalance Handling}\label{moe-imbalance-handling}

The base kernel removes intermediate traffic but does not by itself equalize work. We address imbalance through tile scheduling, execution placement, and communication ordering while preserving the router\textquotesingle s expert choices and the output contract of Section 3.

With prefill-decode disaggregation (PD), the evaluated serving configuration selects EP16 for prefill at \(T \geq 1024\) and symmetric EP8TP2 for decode at \(T \leq 512\), where T is the global token count defined in Section 2.3. This phase-specific layout selection applies only to PD; it is not a general token-count-based switching policy for colocated prefill and decode. The reported decode measurements cover T from 32 to 512.

\subsection{Sources of Imbalance and Cost Models}\label{sources-of-imbalance-and-cost-models}

Uneven routing produces partially occupied tiles, unequal per-device workloads, and bursts of output traffic. For prefill, row or compute-tile counts estimate the work to redistribute. For decode, the number of active experts and their tiles captures weight-streaming costs that token counts alone can miss. These are scheduling proxies, not guarantees of a compute- or bandwidth-bound regime.

Row bucketing and hot-expert interleaving improve local execution (Section 4.2); chipmate redistribution shifts work between devices (Section 4.3); destination interleaving reduces output bursts (Section 4.4). Static placement and preloaded copies address placement and cross-chip skew (Section 4.5), while symmetric TP2 shares decode work within a chip (Section 4.6). These are distinct execution options, not a claim that every mechanism runs in one launch.

\subsection{Load-Aware Tile Scheduling}\label{load-aware-tile-scheduling}

\subsubsection{Row bucketing for partial tiles}\label{row-bucketing-for-partial-tiles}

For uneven expert loads, the kernel selects the smallest precompiled row bucket covering a tile\textquotesingle s valid rows and leading alignment padding, avoiding full-height computation for every tile.

For 41 valid rows at offset eight, the required extent is 49. A 64-row body computes eight leading and fifteen trailing padding rows while skipping half a 128-row allocation. Bucketing reduces but does not eliminate padding, which also remains at block granularity in MegaBlocks {[}11{]}.

\begin{figure}[htbp]
\centering
\includegraphics[width=.95\linewidth]{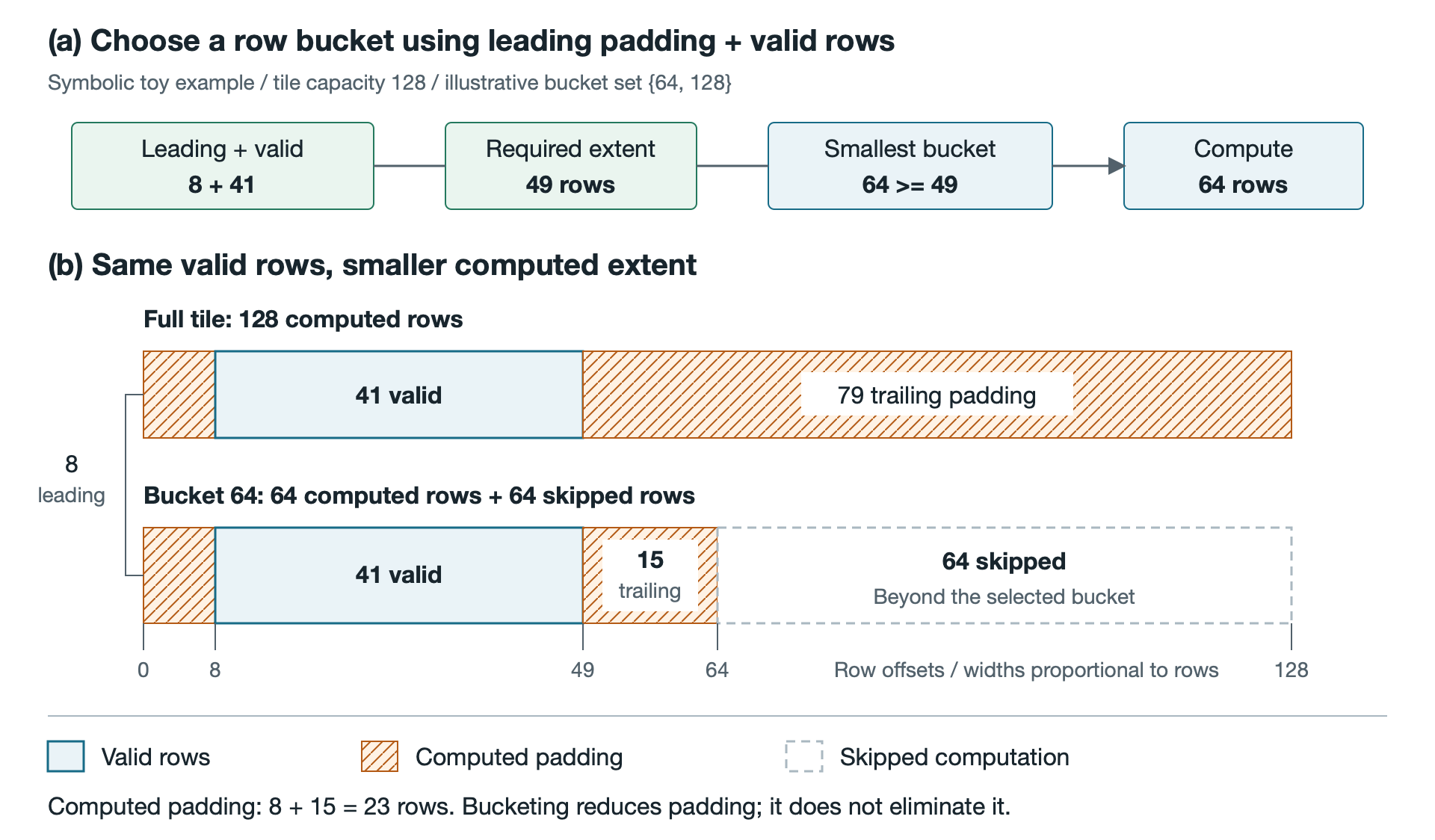}
\caption{ Bucket selection covers valid rows and leading alignment padding, separating computed padding from skipped allocation rows.}
\label{fig:10}
\end{figure}

Each row bucket adds a separately compiled branch of the expert computation. The implementation caps bucket count times the row cost factor for the quantization configuration times \(t_{m}\) at 768, falling back to full-height computation; a 512-row allocation therefore cannot use two buckets. One experiment crossed an instruction-overlay threshold around 66K--70K bundles, triggering code reloads in the loop. This compiler- and shape-dependent observation requires remeasurement and is not total physical instruction-memory capacity.

\subsubsection{Hot-Expert Interleaving}\label{hot-expert-interleaving}

Alternating tiles of the busiest (hot) expert with other experts uses longer hot-expert computation to hide short-lived experts\textquotesingle{} weight loads. A stable merge preserves each stream\textquotesingle s order and leaves unmatched tiles in a tail.

Scheduling uses the maximum, across devices, of busiest-expert load divided by the mean expert load on that device. Counts exclude invalid routes and precede chipmate redistribution. If the hot expert moves, another local expert supplies hot tiles without recomputing the global skew test.

\begin{figure}[htbp]
\centering
\includegraphics[width=.97\linewidth]{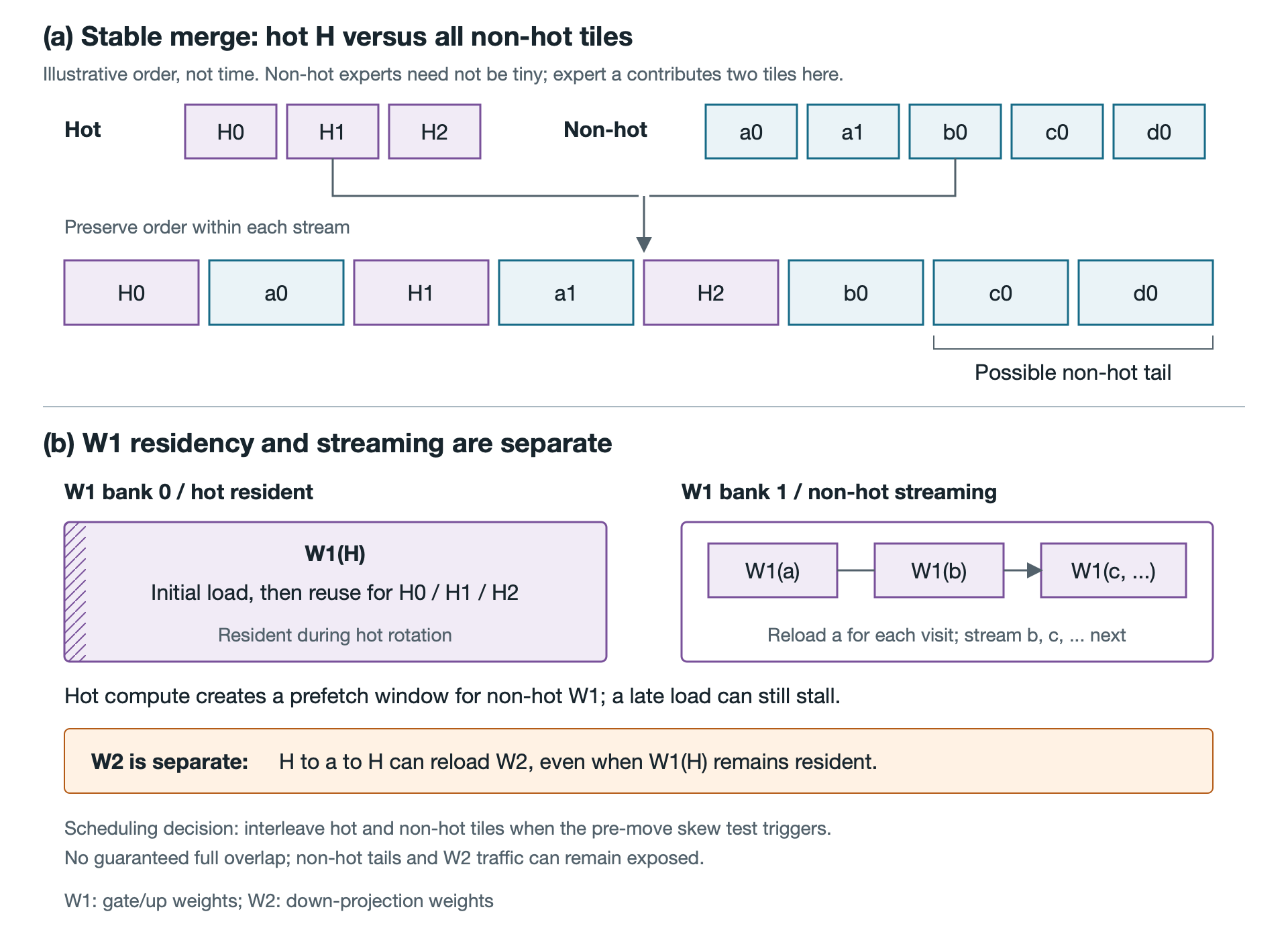}
\caption{ Hot/non-hot interleaving creates prefetch windows. A dedicated W1 bank retains hot weights; W2 follows consecutive-expert residency rules. Unmatched non-hot tiles form a tail, and weight waits remain.}
\label{fig:11}
\end{figure}

Separate VMEM banks retain hot W1 weights while other experts reload theirs on each return. W2 may also reload on expert changes. The added prefetch overlap costs VMEM and weight traffic. Borrowed work remains at the schedule\textquotesingle s end.

\subsection{Intra-Chip Expert Redistribution}\label{intra-chip-expert-redistribution}

Chipmates offer a low-cost rebalance target. Moving an expert\textquotesingle s work copies its weights into receiver HBM and changes the tile schedule for one invocation, not permanent ownership. Inputs are already replicated, so no additional token exchange is needed.

\emph{Planning.} Replicated valid-row counts and the topology map determine at most one expert move within each physical chip\textquotesingle s pair of TensorCore devices per invocation, without a planning collective. Different eligible pairs can move concurrently. The sender removes the selected tiles and the receiver appends them; SMEM holds the move description. A move must improve the chosen load estimate.

\emph{Regime-specific cost.} For large prefill batches, the planner chooses a move that lowers the pair\textquotesingle s maximum row load. Small-batch planning estimates weight cost by active-expert count: it chooses a few-row expert from a globally maximal-count device, requires an active-count gap of at least two, and rejects increases in the pair\textquotesingle s maximum row count. The planner does not explicitly model transfer cost.

These are cost proxies, not regime guarantees. Few rows limit added compute, not weight bytes; improving one tied maximal sender need not reduce the global maximum.

\begin{figure}[htbp]
\centering
\includegraphics[width=.94\linewidth]{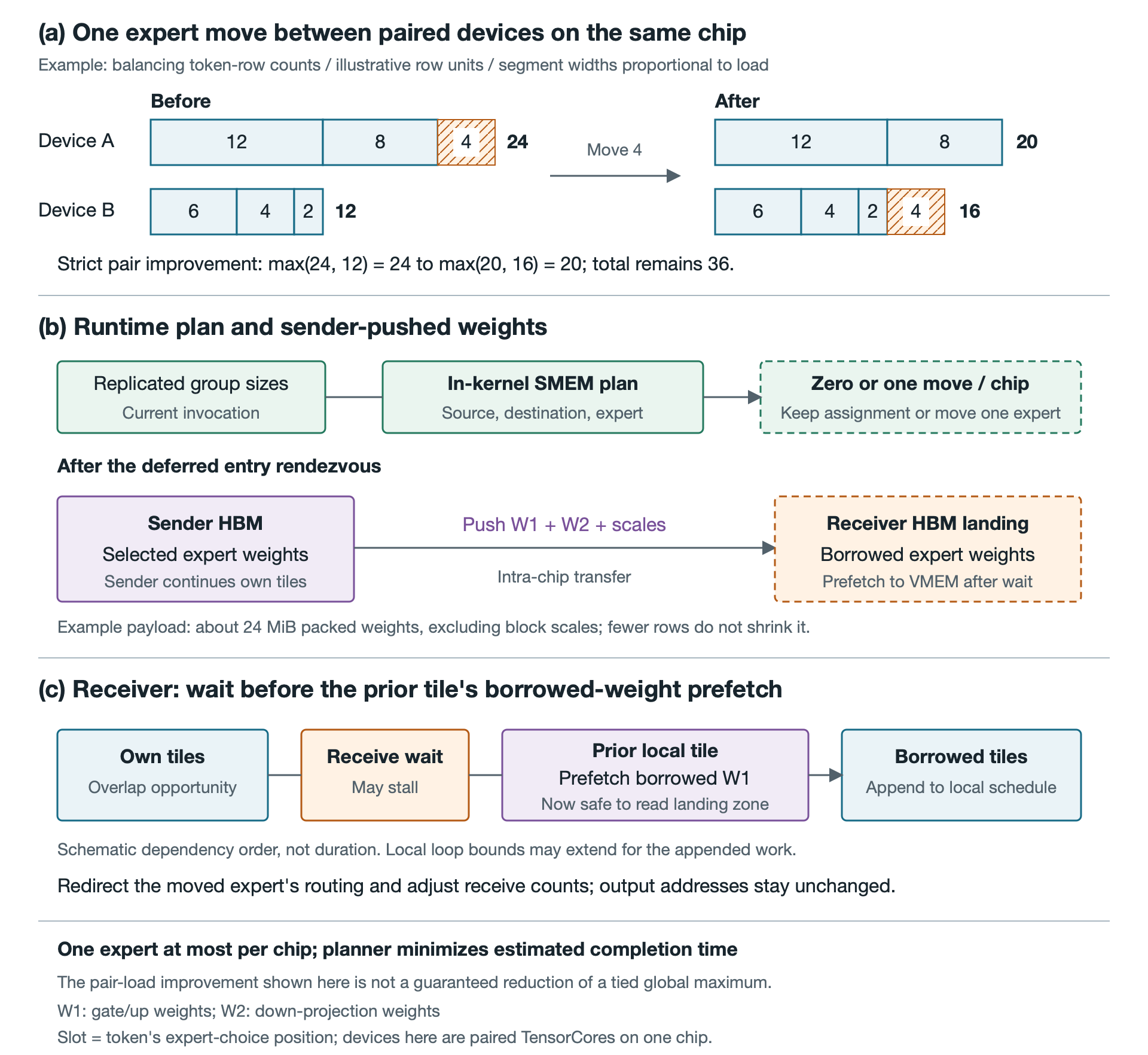}
\caption{ Moving a four-row expert changes example loads from 24/12 to 20/16 but still transfers full W1/W2 and scales. The receiver appends the work and waits before its first weight prefetch.}
\label{fig:12}
\end{figure}

\emph{Weight transfer.} The sender copies gate/up and down weights plus block scales into receiver HBM landing buffers. Local and borrowed weights feed the same matmul body.

Transfer starts after entry readiness and drains at sender exit. The receiver waits before its last local tile prefetches borrowed W1, or at tile 0 if it has no local work. Global scales come from a replicated table prepared outside the kernel.

\emph{Overlap.} The example expert transfers 24 MiB of packed W1/W2, excluding scales. Long resident schedules can hide more of this cost; short schedules may expose most of it. Chip-local transfers still contend for VMEM/HBM bandwidth.

\emph{Routing and completion.} Moving tiles leaves token-slot destinations unchanged but changes local/remote classification and receive counts. The receiver extends its loop bound beyond the shared pre-move bound if needed. Weight transfers start only after entry readiness and drain at exit; idle devices retain these obligations. The exit barrier waits for the slowest peer.

\subsection{Balancing Output Traffic}\label{balancing-output-traffic}

Expert-major, global-token order can create bursts to one destination under contiguous token ownership. Sorting within each expert by (local\_token * P + destination) * k + slot encourages interleaving while preserving expert grouping and original token-slot identities. Here P denotes the number of EP ranks (TensorCore devices), not physical chips.

\begin{figure}[htbp]
\centering
\includegraphics[width=.97\linewidth]{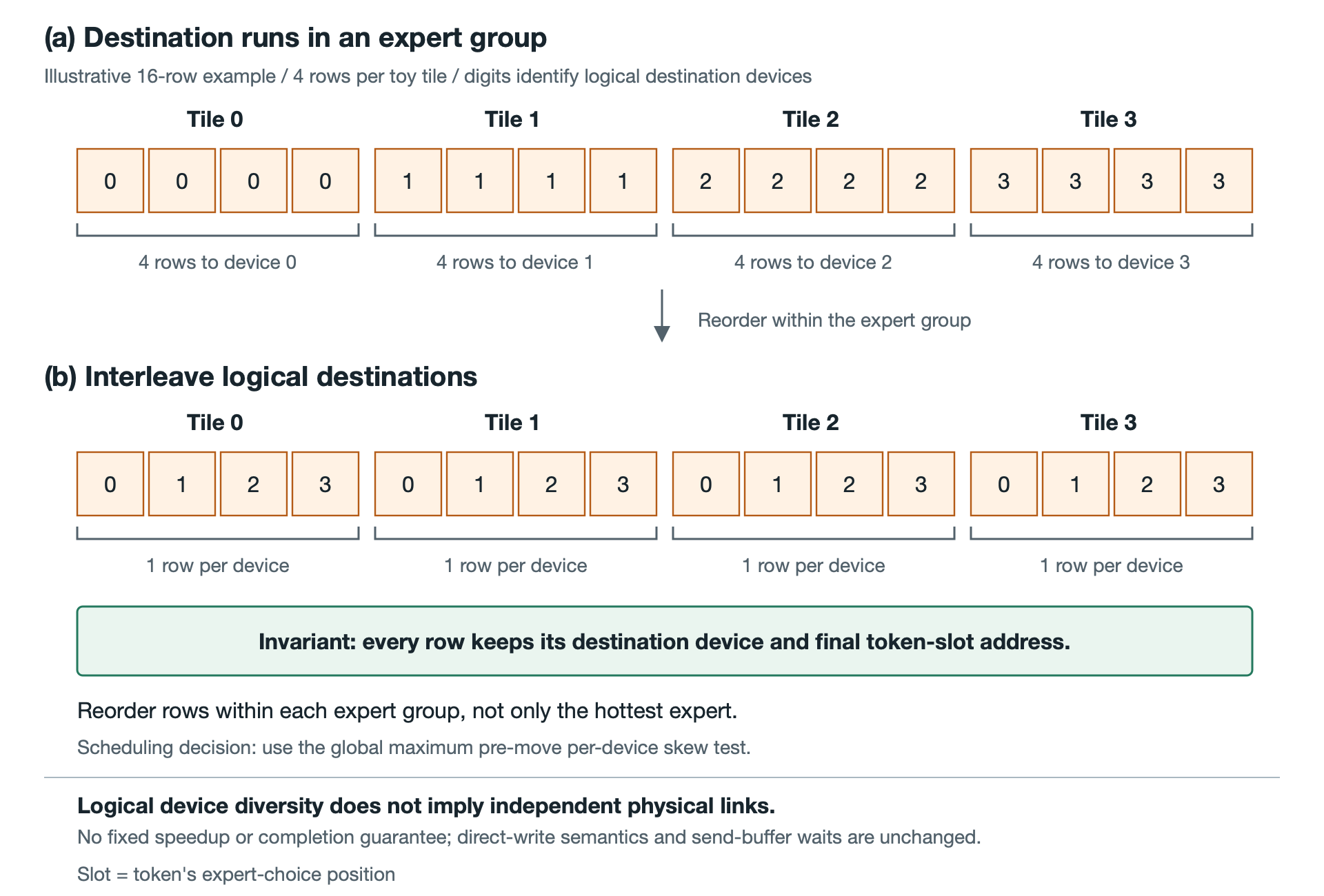}
\caption{ Destination interleaving reduces multi-shard bursts but need not produce round-robin traffic under sparse or skewed routes. Logical destinations share physical links; no P-fold bandwidth gain is implied.}
\label{fig:13}
\end{figure}

The same workload-skew decision controls destination interleaving for all expert groups and hot-expert tile rotation. Noncontiguous destinations require per-row DMA issue (\S{}3.4), trading address work for less bursty traffic. Benefits depend on routing and topology {[}8{]}.

\subsection{Static Placement and Preloaded Expert Copies}\label{static-placement-and-preloaded-expert-copies}

\subsubsection{Static Expert Placement}\label{static-expert-placement}

Chipmate moves cannot change a chip\textquotesingle s total load. Static placement uses observed expert loads to permute router checkpoint columns and expert weights consistently before distribution, retaining fixed expert-slot counts per device. Each expert is assigned once; this is neither replication nor explicit optimization of chip-pair totals.

\subsubsection{Cross-chip balancing with preloaded hot experts}\label{cross-chip-balancing-with-preloaded-hot-experts}

Intra-chip balancing cannot change a chip\textquotesingle s aggregate load. The cross-chip design reserves one additional expert slot per chip. During model weight loading, calibration counts identify hot experts to copy onto other chips. Placement considers the holder\textquotesingle s load and gives each copied expert one non-home holder. The copy and its scales reside in HBM; only working tiles are staged in VMEM. Under TP2, the copied expert is partitioned across the chipmates like the home experts.

At runtime, replicated routing counts determine whether a copied expert executes on its home chip or its holder. This changes execution placement, not the router\textquotesingle s expert choices or weights. A selected expert\textquotesingle s routed work moves as a unit: the home suppresses its tiles and the holder reads the preloaded bank. Token-slot destinations are unchanged, and completion accounting follows the actual writers. No runtime weight transfer or additional input exchange is required.

Figure 14 contrasts the implemented home-or-holder decision with a split-token extension. Splitting would assign disjoint rows of one expert to both chips and requires finer-grained placement and completion accounting; it is not implemented by the current planner.

\begin{figure}[htbp]
\centering
\includegraphics[width=\linewidth]{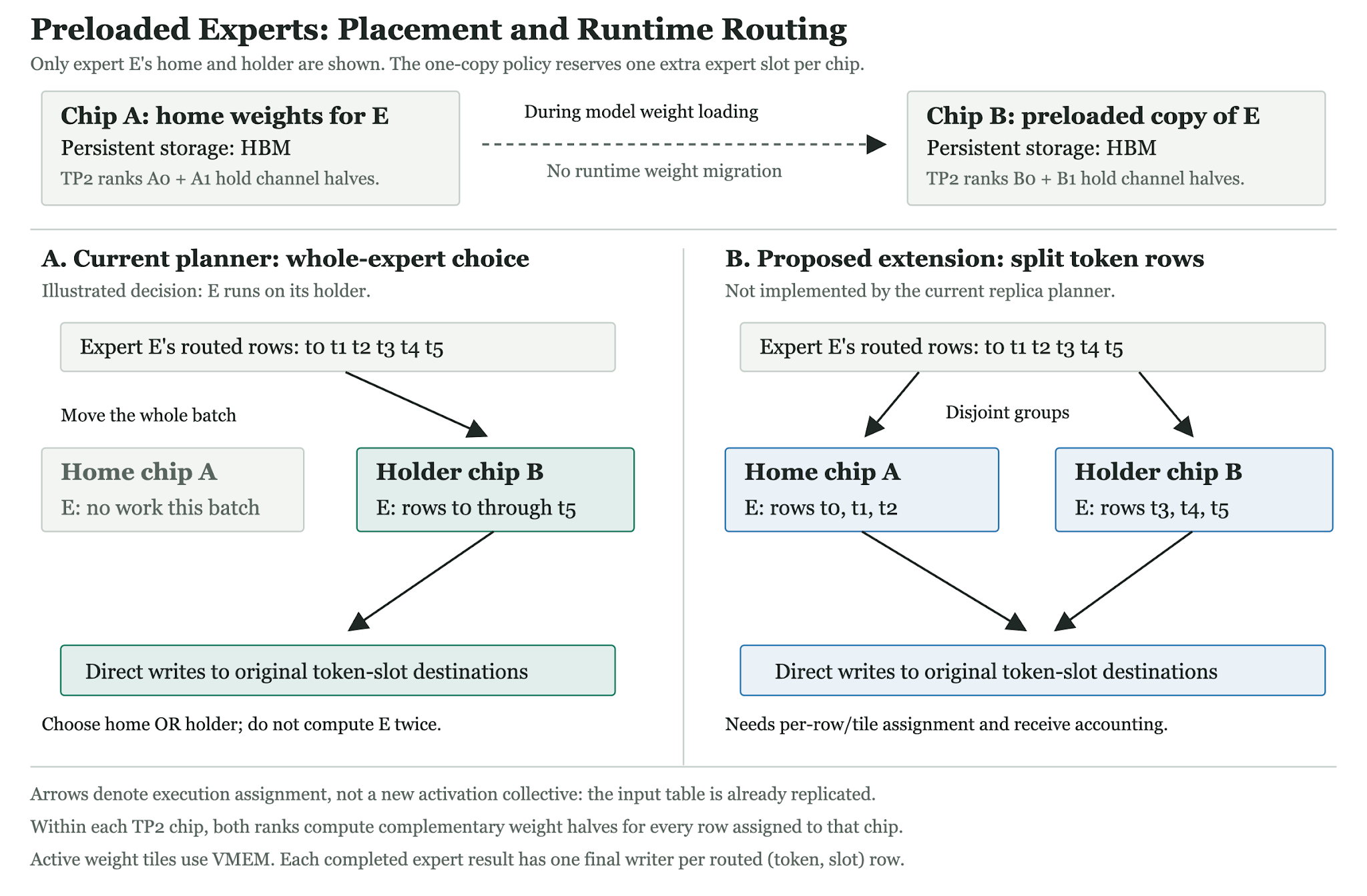}
\caption{\textbf{Preloaded expert routing.} Weights are copied at startup. Left: the current planner assigns the entire expert batch to its home or holder. Right: a proposed extension distributes disjoint token rows across both copies. Arrows denote execution assignment, not an additional activation collective; original token-slot destinations are preserved.}
\label{fig:14}
\end{figure}

The objective depends on the inference regime. A prefill extension would balance per-chip computation, estimated from routed rows or compute tiles. For decode, the number of activated experts captures the cost of fetching many weight matrices for few tokens. The implemented cross-chip decoder uses tile count, which approximates active-expert count for one-tile experts and also accounts for multi-tile experts. For expert cost \(t\) and chip loads \(L_{\mathrm{home}}\), \(L_{\mathrm{holder}}\), it accepts a move when \(L_{\mathrm{home}} - t \geq L_{\mathrm{holder}} + t\), avoiding a transfer that reverses the imbalance.

One copy per chip bounds the added HBM capacity to one expert, including scales. It only helps experts with a pre-positioned copy and does not split a single expert across chips. Replica placement balances aggregate chip load; symmetric TP2 scheduling (Section 4.6) balances execution within the selected chip. Both remain sensitive to routing skew and the slowest chip.

\emph{Implementation scope.} The current optional bank supports a configurable number of copies; the design above is the one-copy configuration. Cross-chip serving is implemented for decode/verification shapes up to 512 tokens and is disabled by default. Applying the same preloaded-copy mechanism to prefill with a row- or compute-based objective is an extension, not a result of the current evaluation.

\subsection{Symmetric EP8TP2 Decode}\label{symmetric-ep8tp2-decode}

\subsubsection{Chip-Local Expert Partitioning}\label{chip-local-expert-partitioning}

Decode cost can be dominated by the number of active experts and weight tiles, even when each expert receives few tokens. EP8 x TP2 shares each chip-owned expert across two cores: W1 is partitioned across matching gate/up channels and W2 across the reduction dimension. Both cores process the same input rows using complementary weight halves, without migrating weights during an invocation.

For input tile X and core p in \{0,1\}:

\[
\begin{aligned}
H_p &= \operatorname{act}(XW_{\mathrm{gate},p}) \odot (XW_{\mathrm{up},p}),\\
P_p &= H_pW_{\mathrm{down},p},\\
Z &= P_0 + P_1.
\end{aligned}
\]

The partial sum reconstructs one expert result before normalization and router weighting. Compared with EP16 fixed ownership, both chipmates participate in every active expert, reducing intra-chip imbalance. Cross-chip skew is addressed separately by placement and preloaded copies (Section 4.5).

\subsubsection{Symmetric Tile Scheduling}\label{symmetric-tile-scheduling}

The symmetric schedule computes the same tile on both cores with identical valid-row bounds and row-bucket selection. Complementary weight partitions therefore carry matched matmul work, avoiding waits caused by unequal tile shapes. This aligns computation without requiring clock-step lockstep. Reduction of tile t is delayed until the next iteration so that partial exchange overlaps later computation. Additional partial buffering and a separate reduction step provide the overlap window.

\subsubsection{Cooperative gather and delayed reduction}\label{cooperative-gather-and-delayed-reduction}

After both cores gather the first tile, subsequent input gathers alternate between cores; the on-chip bridge forwards each gathered tile to its partner. Weight prefetching runs ahead subject to buffer lifetimes. Each core keeps one valid-row half of its output partial and sends the other half, then adds the received contribution for its owned rows and writes those completed rows to their original token-slot destinations. Each routed row therefore has one final writer, although both cores contribute to the expert computation.

TP partials stay in VMEM and use the on-chip bridge rather than an HBM intermediate or a separate TP-reduction launch. FP32 partial storage and transport do not imply full FP32 accumulation: the shared inner kernel retains its BF16 rounding boundaries. Figure 15 shows the dependency schedule.

\subsubsection{Algorithm 2: symmetric decode expert pipeline}\label{algorithm-2-symmetric-decode-expert-pipeline}

Dependency-level outline. Finishing a tile waits for its partner partial, destination metadata, and output entry readiness before reducing and writing owned rows.

\begin{Verbatim}[fontsize=\footnotesize,breaklines=true,breakanywhere=true]
establish chipmate VMEM readiness; build the shared tile schedule
prime weights and inputs; both cores gather tile 0, if present
for tile t:
    gather t+2 and forward t+1 on their assigned cores
    await inputs, weights, and safe buffer reuse; compute P_p(t)
    send the partner-owned partial rows over the bridge
    if t > 0: finish tile t-1
finish the final tile, if any; drain bridge and output DMA
complete entry readiness even when idle; report remote-row counts
rendezvous across EP ranks; wait for incoming output rows
\end{Verbatim}

\begin{figure}[htbp]
\centering
\includegraphics[width=\linewidth]{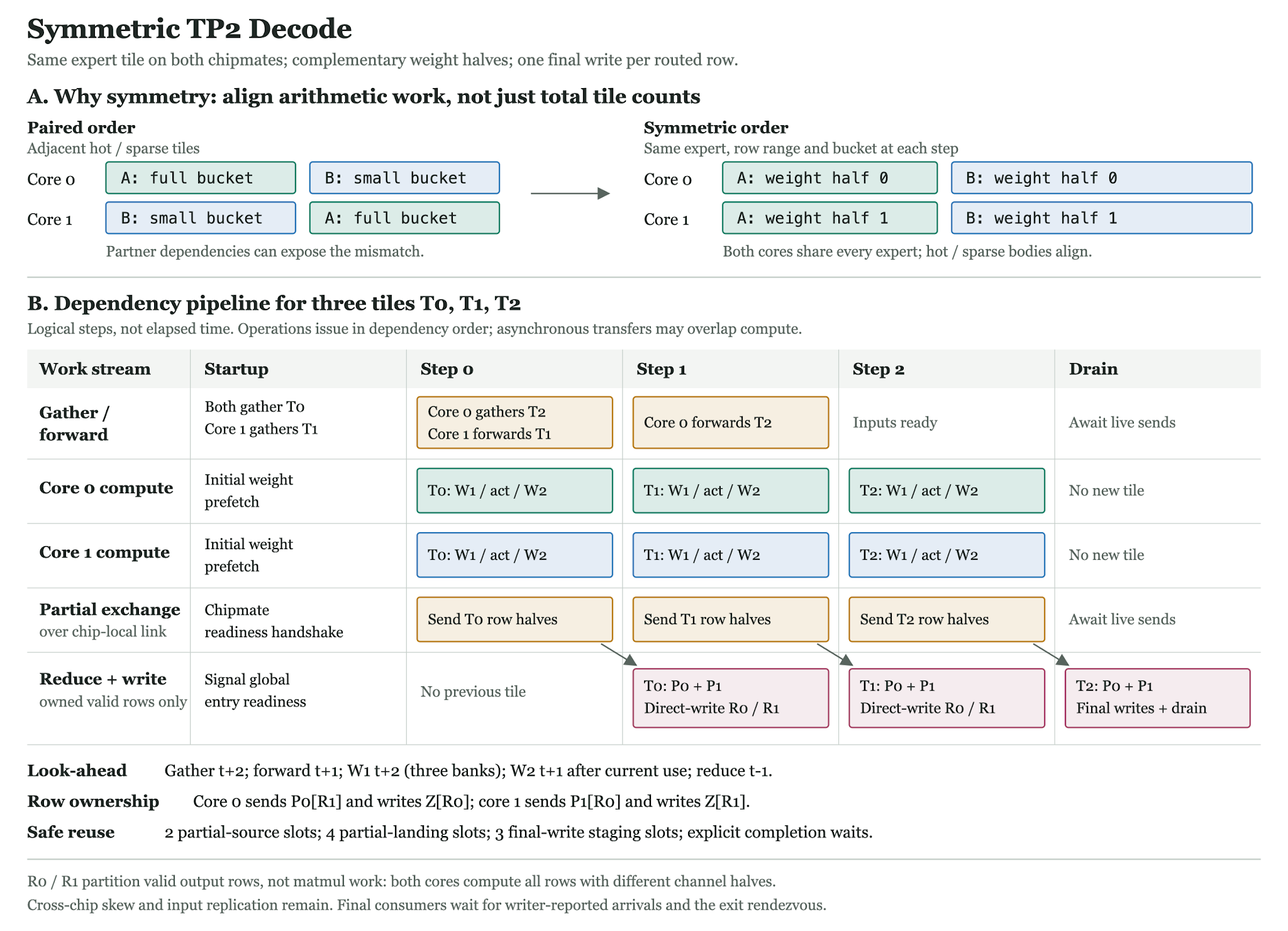}
\caption{ Symmetric TP2 pipeline. Both cores compute the same tile with complementary weight halves; input sharing, weight prefetch, partial exchange, and delayed reduction overlap subject to buffer-lifetime dependencies. The diagram is a dependency schematic, not a measured timeline.}
\label{fig:15}
\end{figure}

\subsubsection{Scope and validation}\label{scope-and-validation}

The evaluated symmetric EP8TP2 path accepts FP4 weights and BF16 inputs for up to 512 global input tokens, with the token count divisible by the participating device count. Section 5.2 reports its kernel-operation latency relative to fused EP16 at 32-512 tokens. Its benefit depends on whether reduced intra-chip imbalance and waiting outweigh bridge and reduction costs.

\FloatBarrier
\section{Empirical Evaluation and Validation Protocol}\label{empirical-evaluation-and-validation-protocol}

\subsection{Model and Measurement Scope}\label{model-and-measurement-scope}

Evaluation uses a TPUv7 (Ironwood) 2x2x2 slice with eight chips and sixteen TensorCore devices. The FP4 model has 256 experts, top-8 routing, and widths d=f=4096. EP16 assigns 16 experts to each device; EP8TP2 assigns 32 experts to each chip, partitioned across its two devices. The Megablox-only EP16 implementation is the baseline for the reported serving comparison; decode kernel-operation results compare symmetric EP8TP2 with our fused EP16 kernel. Scale-block width, full model/deployment configuration, and source revisions remain to be documented.

\begin{table}[!htbp]
\centering
\small
\renewcommand{\arraystretch}{1.12}
\caption{ Model and evaluation hardware. Both EP16 and EP8TP2 use sixteen TensorCore devices on eight TPUv7 chips.}\label{tab:5}
\begin{tabular}{@{}
  >{\raggedright\arraybackslash}p{(\linewidth - 2\tabcolsep) * \real{0.4500}}
  >{\raggedright\arraybackslash}p{(\linewidth - 2\tabcolsep) * \real{0.5500}}@{}}
\toprule
\begin{minipage}[b]{\linewidth}\raggedright
Parameter
\end{minipage} & \begin{minipage}[b]{\linewidth}\raggedright
Value
\end{minipage} \\
\midrule
Hidden / intermediate width & 4096 / 4096 \\
Experts / top-k & 256 / 8 \\
Hardware slice & TPUv7 (Ironwood), 2\ensuremath{\times}2\ensuremath{\times}2 physical chips \\
EP16 devices / experts per device & 16 / 16 \\
W1 / W2 shape per expert & 4096\ensuremath{\times}8192 / 4096\ensuremath{\times}4096 \\
Packed W1 / W2 bytes per expert & 16 MiB / 8 MiB, excluding scales \\
Assignments at T tokens & M = 8T globally; mean M/P per rank under balanced routing \\
Arithmetic configuration & FP4 block-scaled weights; W4A8 or W4A16 must be reported explicitly \\
\bottomrule
\end{tabular}
\end{table}

Report expert-launch, full-MoE-layer, and serving timings separately. Full-layer timing includes preparation, exchange, and combination; distinguish inclusive/exclusive preprocessing costs {[}4{]}. Kernel attribution requires a controlled backend swap.

Precompile and warm up shapes, synchronize asynchronous completion, match inputs/routes and topology, and report repetitions and variability. Prefill, decode, and mixed workloads require valid/padded token counts, concurrency, and route distributions. Uniform routing provides a bound, not representative skew.

\subsection{Performance Results}\label{performance-results}

\textbf{EP16 prefill imbalance.} For the measured skewed prefill workload, imbalance handling reduces kernel runtime from 2.3x to 1.09x the balanced-kernel reference. Both ratios use the same fixed balanced-workload kernel time.

\textbf{Symmetric EP8TP2 decode.} Across decode token counts from 32 to 512, symmetric EP8TP2 reduces kernel-operation latency by up to 21.3\% relative to our fused EP16 kernel. This kernel-operation comparison is distinct from the Megablox-based whole-stack serving result.

\textbf{End-to-end serving.} Relative to the Megablox-based EP16 baseline on the same 2x2x2 slice, whole-stack serving throughput improves by 12-34.5\% across different request mixes and batch sizes. These gains include serving-stack changes and do not isolate the MoE kernel\textquotesingle s contribution.

\subsection{Ablation and Bottleneck Analysis}\label{ablation-and-bottleneck-analysis}

\textbf{FLOPs and traffic.} For \(T\) valid global tokens, top-\(k\) routing, and widths \(d\) and \(f\), the three expert projections perform \(F_{\mathrm{useful}} = 6Tkdf\) FLOPs. Issued matmul work additionally reflects row and width padding; vector work for quantization and activation is separate. With \(A\) active experts and one read of each packed FP4 weight, the weight-only arithmetic intensity is \(I_{\mathrm{W}} = 4Tk/A\) FLOPs/byte. This optimistic intensity excludes scales, activations, metadata, output traffic, and weight reloads.

\begin{table}[!htbp]
\centering
\small
\renewcommand{\arraystretch}{1.12}
\caption{Analytic projection work and weight-only intensity, not measured performance. Balanced routing across all 256 experts, top-8, \(d\) = \(f\) = 4096, and one weight read per expert are assumed. FLOPs cover the entire sixteen-device invocation.}\label{tab:6}
\begin{tabular}{@{}
  >{\raggedright\arraybackslash}p{(\linewidth - 4\tabcolsep) * \real{0.2400}}
  >{\raggedright\arraybackslash}p{(\linewidth - 4\tabcolsep) * \real{0.3800}}
  >{\raggedright\arraybackslash}p{(\linewidth - 4\tabcolsep) * \real{0.3800}}@{}}
\toprule
\begin{minipage}[b]{\linewidth}\raggedright
Global tokens
\end{minipage} & \begin{minipage}[b]{\linewidth}\raggedright
Useful projection GFLOPs
\end{minipage} & \begin{minipage}[b]{\linewidth}\raggedright
Weight-only FLOPs/byte
\end{minipage} \\
\midrule
32 & 25.8 & 4 \\
128 & 103.1 & 16 \\
512 & 412.3 & 64 \\
2048 & 1649.3 & 256 \\
8192 & 6597.1 & 1024 \\
\bottomrule
\end{tabular}
\end{table}

\textbf{Decode.} Few rows per active expert give little weight reuse per FLOP; row-bucket padding and fixed metadata, DMA-issue, and synchronization costs can dominate useful computation. At fixed token count, activating more experts can increase weight traffic without increasing useful FLOPs. Active-expert counts, issued-to-valid row ratios, and the slowest rank therefore complement token count as predictors of latency.

\textbf{Prefill.} More rows per expert amortize weight traffic when the weight panels remain resident across row tiles. If the selected panel set does not fit the buffers, or execution order breaks reuse, repeated loads can limit this gain. Higher token count consequently does not guarantee compute-bound execution: output traffic, exposed staging waits, and skew on the busiest rank can still determine runtime.

\textbf{SOL accounting.} Bounds must use the actual schedule and a consistent per-device scope. An optimistic compute/HBM bound is 
\[
  t \geq \max_r\;\max\left(
    \frac{F_{\mathrm{MXU},r}}{P_{\mathrm{MXU},r}},
    \frac{B_{\mathrm{HBM},r}}{\mathrm{BW}_{\mathrm{HBM},r}}
  \right),
\]
where \(F_{\mathrm{MXU},r}\) counts issued matmul work, not vector operations. Interconnect and synchronization constraints can raise runtime further; overlapping component times must not be added as if serialized. Report useful FLOP/s separately from issued matmul throughput. Total HBM utilization requires measured traffic; a weight-byte model yields effective weight bandwidth, not a hardware-counter utilization reading.

The aggregate results above do not isolate each scheduling and communication mechanism. The following ablations are further validation, not completed results. Each configuration must enter the intended compiled path: small shapes can bypass an optimization even when its flag is enabled.

\begin{itemize}
\tightlist
\item
  \textbf{Fusion boundary.} Measure expert-launch and full-layer latency, including routing, exchange, and postprocessing; obtain baseline launches and materialized bytes from compiled traces.
\item
  \textbf{Weight reuse.} Hold row order and tile shape fixed while retaining or reloading weights; report weight bytes, reuse runs, and layer latency.
\item
  \textbf{Staging depth.} Compare two and three send buffers under a matched memory budget where possible; measure exposed reuse waits and exit tails, not only ICI bytes.
\item
  \textbf{Direct-write issue.} Compare generic and ordered issue on destination-sorted tiles, then separately evaluate interleaving, which removes that ordering condition.
\item
  \textbf{Row specialization.} Compare bucketed and full-row bodies on identical routes; record padded computation, instruction footprint, and overlays.
\item
  \textbf{Load management.} Isolate static placement, hot/non-hot rotation, and chipmate rebalance; report per-rank rows, active experts, transfer bytes, and longest-rank tails.
\item
  \textbf{Preloaded expert copies.} Compare zero and one copy per chip under identical routes and a fixed calibration profile, separate from evaluation traffic. Report per-chip tiles and active experts, slowest-chip latency, added HBM capacity, and startup loading cost. Prefill extensions additionally require compute-load measurements.
\item
  \textbf{Precision and capacity.} Compare FP8/BF16 activations with FP4 weights, accumulator precision, and accuracy; separate numerical changes from scheduling gains.
\end{itemize}

Compute-only and communication-only timings are diagnostic bounds, not additive costs, because they alter dependencies and contention. Measured timelines require profiler data; the figures in Sections 3 and 4 show dependencies only.

\subsection{Model Quality}\label{model-quality}

All production evaluations passed, with no numerical issues observed. This reports production acceptance, not bitwise equivalence between different arithmetic or reduction paths. Detailed evaluation-suite settings and numerical-error summaries remain to be documented. Section 3.4 states the output-ownership and DMA-completion invariants.

\subsection{Production Integration}\label{production-integration}

Reported production use requires a rollout record specifying model coverage, numerical/scheduling settings, integration boundaries, and monitored correctness/performance signals.

\section{Discussion and Future Directions}\label{discussion-and-future-directions}

Fusion avoids intermediate HBM traffic; routing identity and separate source/destination completion make the pipeline safe. Preloaded copies and symmetric TP2 address different levels of imbalance. Scalar overhead, padding, and the slowest-chip tail remain measurement priorities.

\subsection{Upstream Activation Exchange}\label{upstream-activation-exchange}

Consumer-directed upstream exchange could send tokens only to ranks owning selected experts and deduplicate local uses. Savings depend on distinct remote consumers, potentially numerous under top-8 routing.

Per-forward token caching requires invocation isolation and readiness before collection; relay routing requires topology-aware byte/hop and contention analysis. Neither is an established FP4-kernel capability.

\subsection{Adaptive Placement and Prefill Extension}\label{adaptive-placement-and-prefill-extension}

Preloaded copies cover a fixed set of hot experts. Extending cross-chip placement to prefill requires a compute-aware objective and serving integration. Further work includes adapting copy placement as routing changes, supporting multiple holders, and splitting a dominant expert across chips. These extensions must preserve unique token-slot producers and safe weight and output lifetimes across invocations.

\subsection{Precision, Tuning, and Compiler Visibility}\label{precision-tuning-and-compiler-visibility}

Further opportunities include mixed-precision lowering, more accurate memory-budgeted accumulation, instruction-aware specialization, and representative-route tuning. Generated-instruction and DMA-issue visibility would improve reproducibility; the objective remains the validated layer-level tradeoff.

\FloatBarrier

\end{document}